\documentclass[lettersize,journal]{IEEEtran}
\usepackage{amsmath,amsfonts}
\usepackage{bm}
\usepackage{algorithmic}
\usepackage{algorithm}
\usepackage{array}
\usepackage[caption=false,font=normalsize,labelfont=sf,textfont=sf]{subfig}
\usepackage{textcomp}
\usepackage{stfloats}
\usepackage{url}
\usepackage{verbatim}
\usepackage{graphicx}
\usepackage{cite}
\usepackage{hyperref}
\usepackage{url}
\usepackage{color}
\usepackage{colortbl}
\usepackage{array}
\usepackage{multirow}
\usepackage{booktabs}
\usepackage{pifont}
\usepackage{amssymb}
\usepackage{amsmath}
\usepackage{bm}
\usepackage{tikz}
\definecolor{Gray}{gray}{0.93}
\usepackage{listings}
\usepackage{algorithm}
\usepackage{algorithmic}
\usepackage{enumitem}
\usepackage{tcolorbox}
\usepackage{hhline}
\usepackage{caption}

\newcommand{\veritas}{{\textbf{\textsc{Veritas}}}}
\newcommand{\veritaspp}{{\textbf{\textsc{Veritas++}}}}

\definecolor{Gray}{gray}{0.93}
\definecolor{selfblue}{RGB}{65,105,225} 
\definecolor{softblue}{rgb}{0.9, 0.95, 1.0}
\definecolor{lightblue}{RGB}{220,230,255}
\definecolor{deemph}{gray}{0.62}

\newlength\savewidth\newcommand\shline{\noalign{\global\savewidth\arrayrulewidth
  \global\arrayrulewidth 1pt}\hline\noalign{\global\arrayrulewidth\savewidth}}

\begin{document}

\title{\veritaspp: Value-aware On-Policy Distillation for Perception-Enhanced AIGI Detection}

\author{
	Hao Tan,
    Jun Lan,
	Zichang Tan,~\IEEEmembership{Member,~IEEE},
    Ajian Liu,
    Zijian Yu,
    Chuanbiao Song,\\
    Huijia Zhu,
    Weiqiang Wang,
    Jun Wan,~\IEEEmembership{Senior Member,~IEEE},
    Zhen Lei,~\IEEEmembership{Fellow,~IEEE}
\thanks{Hao Tan, Jun Wan and Zhen Lei are with the School of Advanced Interdisciplinary Sciences, University of Chinese Academy of Sciences (UCAS), Beijing 101408, China, and also with the State Key Laboratory of Multimodal Artificial Intelligence Systems (MAIS), Institute of Automation Chinese Academy of Sciences (CASIA), Beijing 100190, China.
Ajian Liu is with the State Key Laboratory of Multimodal Artificial Intelligence Systems (MAIS), Institute of Automation, Chinese Academy of Sciences (CASIA), Beijing 100190, China.
Jun Wan and Zhen Lei are also with the School of Artificial Intelligence, University of Chinese Academy of Sciences (UCAS), Beijing 100049, China (e-mail: \{tanhao2023, jun.wan, zhen.lei\}@ia.ac.cn, ajianliu92@gmail.com).
}
\thanks{Jun Lan, Zijian Yu, Chuanbiao Song, Huijia Zhu and Weiqiang Wang are with Ant Group (e-mail: \{yuzijian.yzj, songchuanbiao.scb, yelan.lj, huijia.zhj, weiqiang.wwq\}@antgroup.com).}
\thanks{Zichang Tan is with the Sangfor Technologies Inc. (e-mail: tanzichang@foxmail.com).}
}


\maketitle

\begin{abstract}
The growing capability of image generation models has made synthetic images a routine presence in open media, making robust and generalizable AI-Generated Image (AIGI) detection increasingly essential.
While multi-modal large language models (MLLMs) offer a transparent alternative to black-box binary scoring, we observe that current MLLM-based detectors still exhibit notable \textit{perception bottlenecks} in capturing fine-grained anomalies.
They primarily focus on how visual evidence is organized and synthesized, leaving the intrinsic perception less optimized.
To mitigate this gap, we present \veritaspp, a perception-enhanced reasoning framework that establishes reliable perception as the foundation of authenticity reasoning.
Rather than directly optimizing the model's explanatory ability, we ground AIGI detection on three basic perception abilities, i.e., capturing fine-grained visual details, semantic anomalies and pixel-level differences.
Building on this insight, we introduce Perception-oriented Learning (PoRL), which replaces open-ended description supervision with verifiable rewards to explicitly strengthen these capacities.
To further integrate enhanced perception with reasoning, we introduce Value-aware On-Policy Distillation (VaOPD), an adaptive distillation mechanism that prioritizes high-value distillation signals over uniform supervision, internalizing perception-aware reasoning through a privileged self-teacher.
Extensive experiments across standard, in-the-wild and emerging benchmarks demonstrate that \veritaspp$\,$ achieves promising generalization.
The perception learning effectively bridges the perception gap and yields seamless gains on detection, while VaOPD further enables efficient capability evolvement without sacrificing existing performance.
Code and checkpoints are available at \href{https://github.com/EricTan7/VeritasPP}{github.com/EricTan7/VeritasPP}.
\end{abstract}


\section{Introduction}
\label{sec:intro}

\IEEEPARstart{T}{he} rapid evolution of generative models has made high-fidelity synthetic images a routine presence in open visual media.
While such progress empowers creative applications, it also lowers the cost of producing convincing forgeries across diverse semantic scenes, raising persistent concerns about and digital deception.
Consequently, robust and generalizable AI-Generated Image (AIGI) detection has become increasingly crucial for maintaining digital trust and preventing the misuse of synthetic media.
To meet this demand, recent works~\cite{zhou2025aigi,wen2026spot,tan2025veritas} increasingly turn to Multi-modal Large Language Models (MLLMs), whose natural-language outputs offer a transparent alternative to black-box binary scoring~\cite{ojha2023towards, yan2025orthogonal, chen2025dual}.

\begin{figure*}[t]
    \centering
    \includegraphics[width=0.99\linewidth]{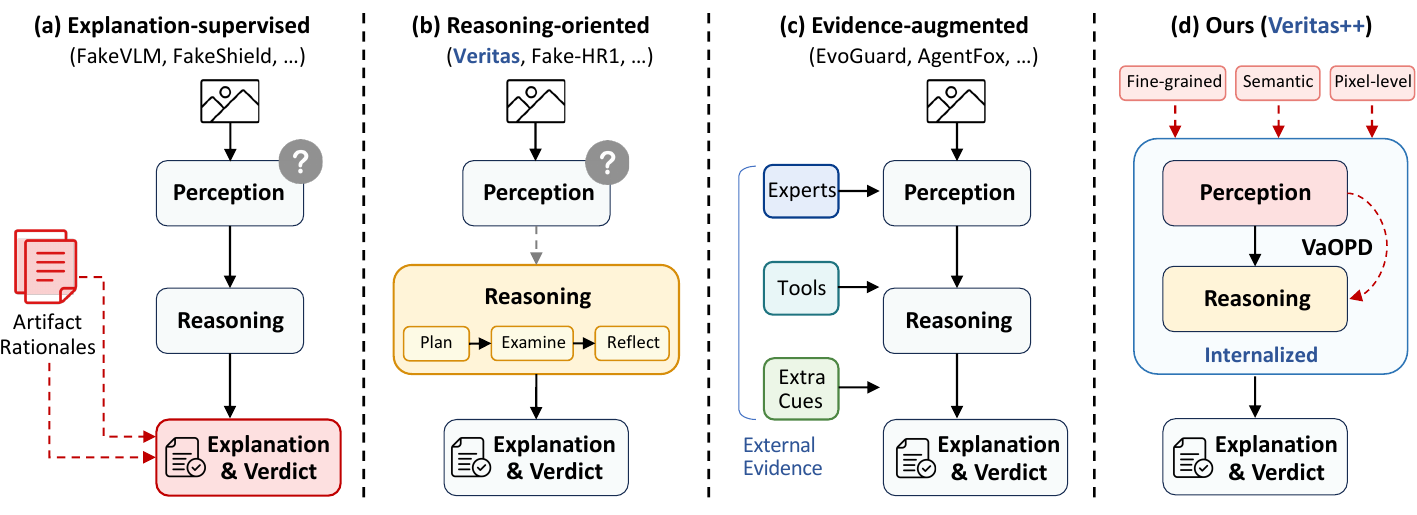}
    \vspace{-4pt}
     \caption{\textbf{Paradigms of MLLM-based AIGI detection.}
     \textbf{(a)} Explanation-supervised methods learn from artifact rationales.
     \textbf{(b)} Reasoning-oriented methods, including \veritas~\cite{tan2025veritas}, organize visual evidence into structured reasoning.
     \textbf{(c)} Evidence-augmented methods incorporate external experts, tools, or cues.
     \textbf{(d)} \veritaspp explicitly strengthens fine-grained, semantic, and pixel-level perception, and internalizes it into reasoning through VaOPD.}
	\label{fig:paradigm}
    \vspace{-0.4cm}
\end{figure*}

Existing MLLM-based AIGI detectors can be roughly organized into three categories, as illustrated in Figure~\ref{fig:paradigm}.
\textbf{(a)} \textit{Explanation-supervised} methods~\cite{wen2026spot,zhou2025aigi,ji2026fakexplain,xu2025fakeshield,huang2025sida,kang2025legion} train MLLMs to generate artifact descriptions.
While these methods improve interpretability, their training objectives are largely centered on producing explanatory rationales, leaving the foundational perceptual abilities required for such explanations less explicitly optimized.
\textbf{(b)} \textit{Reasoning-oriented} methods~\cite{jiang2026fake,zhu2026fakevlm,wen2025busterx++,tan2025veritas,li2026vigil,huang2026realign} move forwards by designing Chain-of-Thought (CoT) outputs, e.g., hybrid thinking~\cite{jiang2026fake} and pattern-aware reasoning~\cite{tan2025veritas}.
However, they primarily focus on specific reasoning formulation, with the effectiveness fundamentally hinging on the quality of the model's perception.
\textbf{(c)} \textit{Evidence-augmented} methods~\cite{cao2025reveal,zhu2026evoguard,yu2026agentfox,ji2026locate,zhang2026unigendet} expand the evidence source through agentic tool calling~\cite{zhu2026evoguard,yu2026agentfox,ji2026locate} or generation feedback~\cite{zhang2026unigendet}.
However, the gains mainly stem from external modules or privileged pipelines,
which can hardly translate into reusable perception ability inside the MLLM.
Although these methods have advanced the interpretability of AIGI detection,  they primarily focus on how evidence is organized and synthesized, leaving the intrinsic perception less optimized.

\begin{figure}[t]
    \centering
    \includegraphics[width=0.99\linewidth]{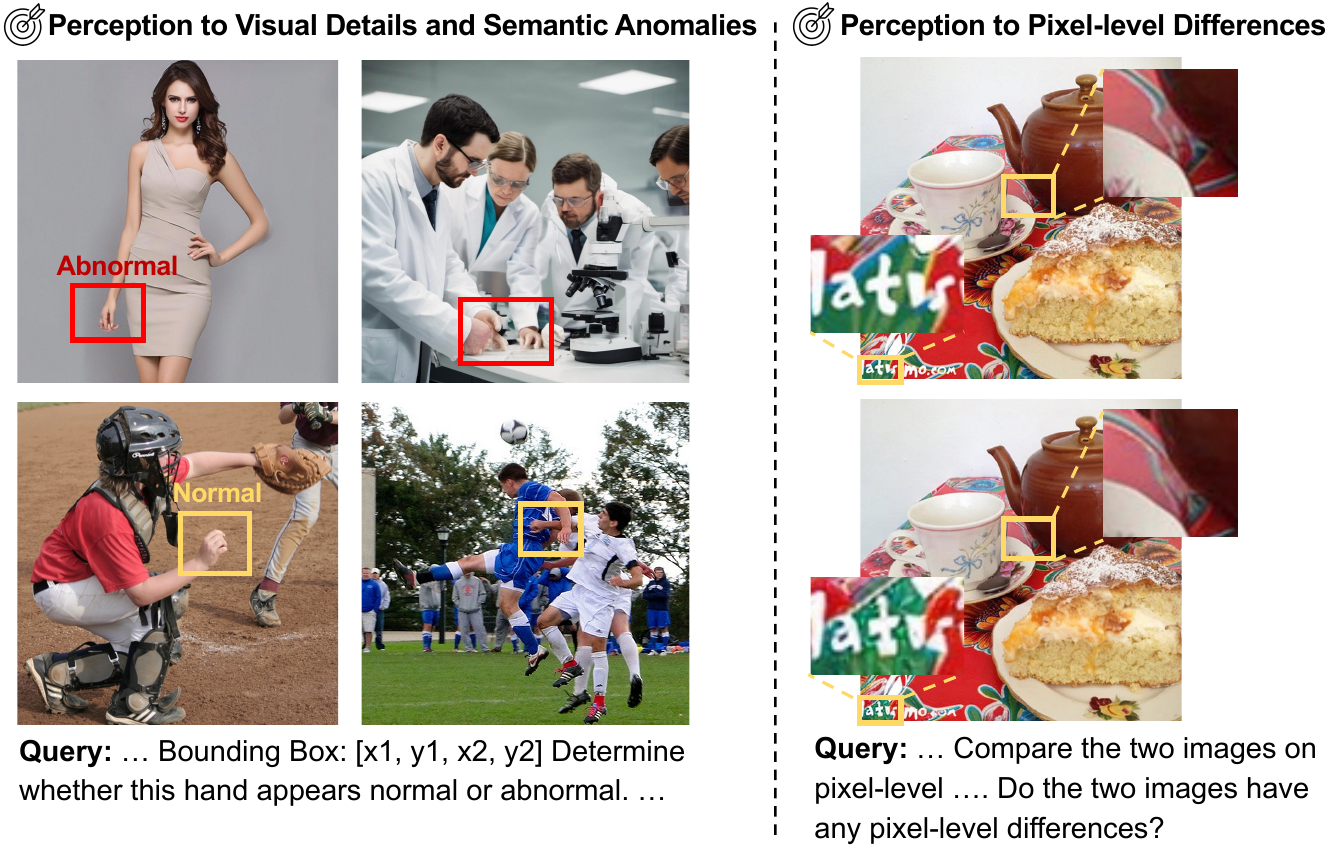}
    \vspace{-4pt}
     \caption{\textbf{Examples of perception challenges in AIGI detection.}
     \textbf{(a)} Fine-grained details and semantic anomalies require localized structural understanding.
     \textbf{(b)} Pixel-level discrepancies can remain subtle even between paired images.}
	\label{fig:perception_example}
    \vspace{-0.4cm}
\end{figure}

Specifically, we observe perception bottlenecks in our preliminary work (i.e., \veritas~\cite{tan2025veritas}) and existing MLLM-based methods.
AIGI detection inherently relies on the capability to capture fine-grained visual details, semantic anomalies and pixel-level discrepancies.
These abilities are broadly relevant to diverse forgeries, e.g., suspicious regions may occur in tiny areas, semantic anomalies may arise from structural deformities (e.g., distorted limbs) or logical contradictions, and low-level artifacts may remain informative even when high-level semantics appear visually coherent.
However, our empirical analysis reveals that existing MLLMs, even those fine-tuned for AIGI detection, exhibit notable \textit{deficiencies} in these areas.
As shown in Figure~\ref{fig:perception_example} and Table~\ref{tab:perception_gap}, 
we evaluate different models on perceiving anatomical and structural flaws (i.e., Abhuman~\cite{fang2024humanrefiner}, MagicData~\cite{wang2025magicmirror} and SDG~\cite{zhang2026and}) and pixel-level differences (i.e., B-Free~\cite{guillaro2025bias}).
The results show that:
(1) Current MLLM-based detectors still exhibit clear perceptual limitations.
(2) Generic MLLMs such as Qwen3-VL are almost incapable of recognizing visual anomalies and pixel-level differences.
(3) Detection fine-tuning alone could not effectively mitigate this gap.

Building upon this insight, we present \veritaspp, a perception-enhanced reasoning framework for AIGI detection.
Our preliminary work, \veritas~\cite{tan2025veritas}, demonstrates that pattern-aware reasoning improves out-of-distribution (OOD) generalization for face forgery detection, primarily addressing \textit{how to reason}.
In this paper, we extend that line from face forgery to general AI-generated image detection, and further ask a complementary question: \textit{how can the model perceive better for more effective reasoning}?
To this end, we introduce improvements from three perspectives over \veritas:
(1) We emphasize high-quality cold start by collecting data only with human-annotated artifacts and rewriting them into a pattern-aware reasoning style.
This simple design fundamentally ensures a low-hallucination and high-quality cold start, which avoids injecting erroneous perception signals.
(2) We introduce Perception-oriented Learning (PoRL), which leverages verifiable rewards to strengthen the model's ability to capture fine-grained visual details, semantic anomalies and pixel-level differences.
As shown in Table~\ref{tab:perception_gap}, this significantly bridges the perception gap, leading to \textit{seamless} improvements in detection performance.
(3) To further integrate the enhanced perception capability with reasoning, we propose \textbf{VaOPD} (Value-aware On-Policy Distillation), which distills perception-aware reasoning back into the model.
Through identifying high-value tokens and adaptive distillation directions, VaOPD greatly improves the learning efficiency, which also enables capability evolution for novel generators and target scenarios.

\begin{table}[t]
    \scriptsize
    \centering
    \caption{Performance comparison on semantic anomaly and AIGC detection benchmarks.}
    \vspace{-4pt}
    \label{tab:perception_gap}
    \renewcommand{\arraystretch}{1.06}
    \scalebox{1.07}{
        \begin{tabular}{
                p{62pt}<{\raggedright}
                p{13pt}<{\centering}
                p{13pt}<{\centering}
                p{14pt}<{\centering}
                p{14pt}<{\centering}
                >{\columncolor{blue!5}}p{13pt}<{\centering}
                >{\columncolor{gray!15}}p{20pt}<{\centering}}
                
                \multirow{2}{*}{\hspace{-5pt}\textbf{Method}}
                & \multicolumn{5}{c}{\textbf{Perception Tasks}}
                &  \\
                \hhline{~-----~}
                & \rule{0pt}{7pt}AbH. & SDG. & Magic.
                & BFree
                & \textbf{Avg.} & \multirow{-2}{*}{\kern-3pt\textbf{Detection}} \\
                \shline
                \hspace{-5pt}Qwen3-VL-8B~\cite{bai2025qwen3}
                & 50.0 & 50.3 & 57.4 & 54.7 & 53.1 & 70.3 \\
                \hspace{-5pt}UniGenDet~\cite{zhang2026unigendet}
                & 50.1 & 55.8 & 52.0 & 50.0 & 52.0 & 70.5 \\
                \hspace{-5pt}FakeVLM~\cite{wen2026spot}
                & 76.6 & 85.2 & 87.4 & 40.2 & 72.4 & 71.2 \\
                \hspace{-5pt}DeepVRM~\cite{lin2026deep}
                & 84.3 & 88.5 & 92.9 & 81.8 & 86.9 & 76.7 \\
                \hhline{-------}
                \hspace{-5pt}\rule{0pt}{6pt}\veritas~\cite{tan2025veritas}
                & 62.7 & 69.1 & 70.2 & 56.4 & 64.6 & 71.5 \\
                \hspace{-5pt}\veritaspp$\,$(SFT)
                & 86.9 & 68.5 & 66.3 & 84.0 & 76.4 & 82.2 \\
                \hspace{-5pt}\veritaspp$\,$(Percept.) & \textbf{97.4} & \textbf{97.5} & \textbf{99.4} & \textbf{92.0} & \textbf{96.6} & \textbf{86.9} \\
        \end{tabular}
    }
    \vspace{-0.4cm}
\end{table}

Our contributions are summarized as follows:
\begin{itemize}
    \item We propose \veritaspp, an interpretable AIGI detection framework that extends the scope of \veritas$\,$ from face forgery to general image detection. Combing enhanced visual perception with pattern-aware reasoning, \veritaspp$\,$ serves as a robust and transparent detector.
    \item We observe perception bottlenecks in existing MLLM-based methods and introduce Perception-oriented Learning (PoRL). This explicitly strengthens the model's capability to perceive fine-grained visual details and anomalies, which fundamentally improves detection.
    \item We propose \textit{VaOPD}, an adaptive self-distillation mechanism that internalizes perception-aware reasoning. This also enables efficient capability evolution for novel generators and target scenarios.
\end{itemize}

\section{Related Work}
\label{sec:related_work}

\subsection{AI-Generated Image Detection}

\noindent \textbf{Task and Datasets.}
AI-Generated Image (AIGI) detection aims to distinguish synthetic images from authentic ones.
This task is becoming increasingly challenging as the image generators evolve rapidly and the forgeries move toward new semantic categories~\cite{zhang2026aegis, qu2026textshield}, application scenarios~\cite{yan2026fraudbench} and real-world degradations~\cite{liu2025beyond, li2025bridging}.
Classic datasets~\cite{zhu2023genimage, hong2025wildfake, ojha2023towards, yan2024df40, park2025community, du2026forensichub} such as GenImage~\cite{zhu2023genimage} provide large-scale evaluations across common generators and authentic images.
More recent datasets~\cite{yan2025sanity, huang2025so, li2026artificial, wen2025busterx, wen2025busterx++, tan2025veritas, zhang2025d3qe, mu2025no, li2025next} such as Chameleon~\cite{yan2025sanity} and HydraFake~\cite{tan2025veritas} emphasize photorealistic synthesis and diverse generation pipelines, posing greater challenges for detectors.
Meanwhile, researchers have attended to practical challenges, incorporating real-world degradations~\cite{liu2025beyond, li2025bridging} to enable more realistic assessments under deployment-oriented conditions.
Emerging benchmarks further target new generators (e.g., GPT-Image-2~\cite{zewde2026gpt}) and specific application scenarios such as text-rich scenes~\cite{qu2026textshield}, e-commerce fraud~\cite{yan2026fraudbench} and academic images~\cite{zhang2026aegis}.
This progression reveals that generation quality and forgery scenarios are continuously evolving, underscoring the need for detectors that can continuously adapt to novel generators and scenarios.

\noindent \textbf{Vision Models for AI-Generated Image Detection.}
AI-generated image detection has been extensively studied with vision-only models.
Early detectors exploit spatial artifacts~\cite{ojha2023towards, yang2025d}, frequency statistics~\cite{frank2020leveraging, qian2020thinking, tan2024frequency} and patch-level inconsistencies~\cite{zhong2023patchcraft, mu2025no, yang2025all}, achieving promising performance within certain generators.
With improving generation quality, recent studies have moved toward more generalizable designs, such as bias-free training~\cite{chen2025dual, rajanaligned, li2026reduce}, semantic decoupling~\cite{yan2025orthogonal, cheng2025co, guo2025omniaid} and dynamic decision boundaries~\cite{shi2026hydraprompt}.
Some methods also explored foundation-model representations~\cite{zhou2026simplicity, liu2026mirror} and incremental adaptation~\cite{hu2026saido, wang2026generalizable}.
These methods reveal that reliable detection depends on both low-level traces and semantic robustness.
However, most vision-based detectors still return a binary score, with limited ability to explain how the evidence is perceived and organized for the judgment.

\noindent \textbf{MLLMs for AI-Generated Image Detection.}
Moving beyond black-box classification, recent methods utilize MLLMs to make transparent detection.
These methods can be categorized into three clusters:
\textbf{(1)} Explanation-supervised methods~\cite{xu2025fakeshield, huang2025sida, wen2026spot, kang2025legion, zhou2025aigi, ji2026fakexplain}, which train MLLMs to produce artifact descriptions or grounded evidence.
For example, FakeVLM~\cite{wen2026spot} builds clue annotations for synthetic image detection, AIGI-Holmes~\cite{zhou2025aigi} aligns explanations through preference data and FakeXplain~\cite{ji2026fakexplain} encourages grounded explanations for visible artifacts.
While these methods focus on interpretability, they leave the foundational perceptual abilities less explicitly optimized.
\textbf{(2)} Reasoning-oriented methods~\cite{jiang2026fake, zhu2026fakevlm, wen2025busterx++, tan2025veritas, li2026vigil, huang2026realign} aim to improve generalization through deep reasoning.
Our conference version, \veritas~\cite{tan2025veritas}, introduces pattern-aware reasoning into deepfake detection, where MLLMs are guided to follow structured thinking patterns such as planning and self-reflection before reaching authenticity verdicts.
Building on the reasoning paradigm,
VIGIL~\cite{li2026vigil} further enforces region-grounded forensics by structuring the inference into a plan-then-examine pipeline, while ReAlign~\cite{huang2026realign} distills reasoning texts into lightweight representations, extending reasoning-oriented detection with additional evidence organization or representation transfer.
While these methods concentrate on particular reasoning strategies, their effectiveness fundamentally depends on the model's perception.
\textbf{(3)} Evidence-augmented methods~\cite{cao2025reveal, zhu2026evoguard, yu2026agentfox, ji2026locate} address this issue by incorporating external evidence.
REVEAL~\cite{cao2025reveal} constructs explicit chains of evidence from additional inputs (e.g., frequency and edge maps), while it forces the model to learn such evidence based on original inputs, making it prone to textual fitting rather than internalizing transferable perception.
Recent methods turn to agentic frameworks~\cite{zhu2026evoguard, yu2026agentfox, ji2026locate}, curating specialized tool sets and agent orchestration to achieve multi-turn evidence synthesis.
However, they heavily rely on external modules or privileged pipelines, which may not be internalized as the MLLM's own capabilities.
Instead of relying on complex scaffolds, we consider visual perception as a fundamental capability beyond reasoning, aiming to strengthen fine-grained visual abilities that support authenticity judgment.

\subsection{Perception and Reasoning in MLLMs}

Recent MLLM studies have paid increasing attention to both perception and reasoning abilities.
Despite their strong multimodal reasoning capabilities, MLLMs have been shown to suffer from fundamental limitations in visual perception~\cite{shi2026vlms, asadi2026mirage}.
Therefore, beyond scaling textual reasoning chains, recent works enhance MLLMs either by active visual interaction or training-time perception internalization.
\textbf{(1)} Agentic methods~\cite{su2026pixel, zheng2026deepeyes, lai2026minio, zhang2026thyme} utilize ``thinking with images'' pipelines to enable active visual perception, e.g., DeepEyes~\cite{zheng2026deepeyes} and Thyme~\cite{zhang2026thyme} equip models with executable image-processing operations (e.g., zooming and visual search),
demonstrating that active inspection can improve difficult multimodal tasks.
However, additional tool calls and multi-turn visual encoding impose heavy deployment cost.
\textbf{(2)} Training-based methods~\cite{huang2026spotlight, wang2026perceptionaware, wei2026zooming, zeng2026agentic} instead try to internalize perception into the model.
ZwZ~\cite{wei2026zooming} distills regional supervision back to full-image understanding, greatly enhancing the fine-grained perception capabilities.
While PAPO~\cite{wang2026perceptionaware} achieves this through algorithmic improvement, enforcing visually grounded reasoning by maximizing the divergence between original and disturbed visual inputs.
These studies underscore the importance of visual perception for robust multimodal reasoning.
Nevertheless, AIGI detection poses a more critical perceptual challenge.
Besides fine-grained semantic understanding, it also requires the ability to capture semantic anomalies and pixel-level irregularities.
In this paper, we treat such capabilities as an important basis for robust reasoning and detection.

\subsection{On-Policy Self-Distillation}

On-Policy Distillation (OPD)~\cite{gu2024minillm, agarwal2024onpolicy} has emerged as an effective fine-tuning technique, where the student receives dense token-level feedback on its own trajectories,
alleviating the exposure bias of off-policy distillation.
To further reduce training cost, On-Policy Self Distillation (OPSD)~\cite{shenfeld2026self, zhao2026self} leverages privileged information for knowledge injection, achieving promising results without requiring external teachers.
Recent variants~\cite{yuan2026vision, yang2026self, jia2026asymmetric, jin2026entropy, lu2026self} improve this by identifying critical tokens during distillation.
Building on its advantages, we adopt OPSD to seamlessly convert the model's perceptual abilities into explicit reasoning ability.
However, existing methods remain suboptimal in learning novel forgeries (e.g., emerging generative models and semantic scenarios).
In this paper, we introduce three-level adaptive reweighting to emphasize tokens with higher distillation value, enabling capability evolvement while preserving existing detection ability.

\section{Method}
\label{sec:method}

\begin{figure*}[t]
    \centering
    \includegraphics[width=0.95\linewidth]{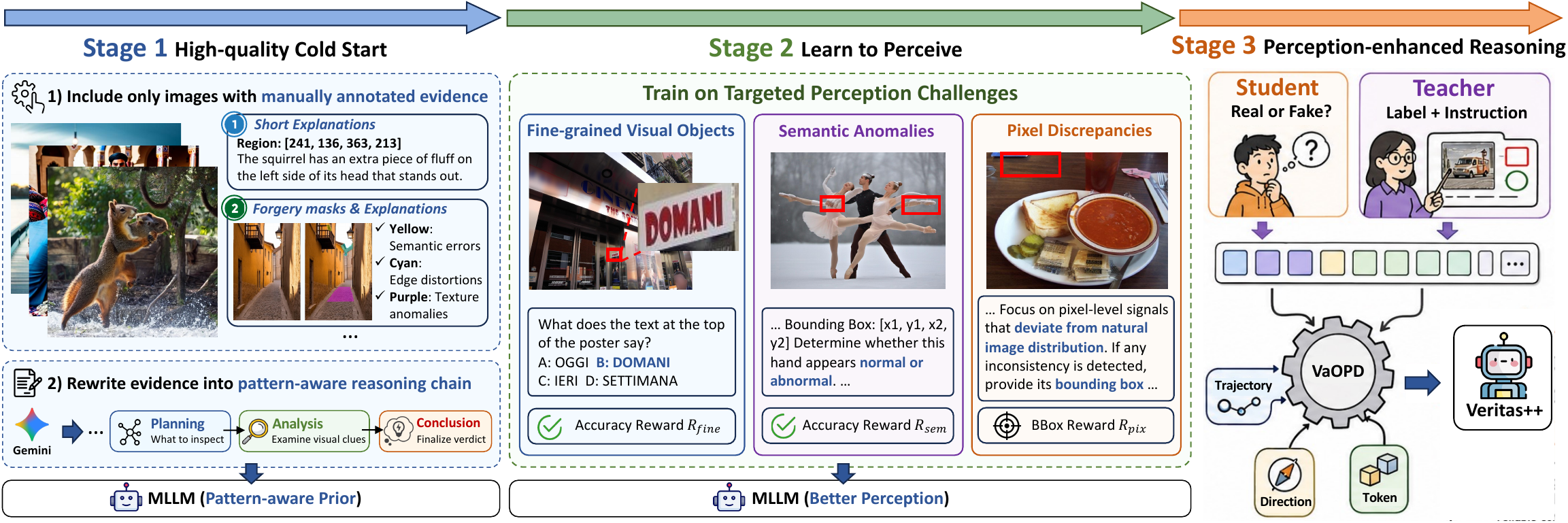}
    \vspace{-5pt}
     \caption{\textbf{Overview of \veritaspp.}
     \textbf{Stage 1: High-quality Cold Start} establishes a reliable, low-hallucination reasoning prior. We retain only images with manually annotated evidence and rewrite their short explanations or forgery masks into pattern-aware reasoning chains.
     \textbf{Stage 2: Learn to Perceive} bridges the perception gap through targeted challenges on fine-grained visual objects, semantic anomalies and pixel-level discrepancies, supervised by verifiable accuracy or bounding-box rewards.
     \textbf{Stage 3: Perception-enhanced Reasoning} internalizes the improved perception into authenticity reasoning. A privileged self-teacher provides dense guidance on student rollouts, while VaOPD adapts the distillation signals at the trajectory, token and direction levels.}
	\label{fig:method}
    \vspace{-0.4cm}
\end{figure*}

In this section, we present the details of \veritaspp.
Building upon our preliminary version \veritas~\cite{tan2025veritas}, which demonstrates that pattern-aware reasoning improves out-of-distribution (OOD) generalization, we extend the scope from facial images to general AI-generated image detection, and further explore the synergy of reliable perception and pattern-aware reasoning.
As shown in Figure~\ref{fig:method}, our framework follows a three-stage training pipeline:
\textbf{(1)} hiqh-quality cold-start that avoids injecting erroneous perception priors,
\textbf{(2)} perception-oriented learning that strengthens multiple levels of perceptual capacities through verifiable rewards,
and \textbf{(3)} Value-aware On-Policy Distillation (VaOPD) that internalizes perception-aware reasoning back into the model and supports capability evolution for novel forgeries.

\subsection{High-quality Cold-Start}
\label{sec:method_cold}

Different from recent practices that harvest artifact descriptions from MLLMs~\cite{wen2026spot,huang2025sida,tan2025veritas}, we observe that machine-generated rationales inevitably introduce perception hallucinations (e.g., describing non-existent artifacts), which contaminate the cold-start signal.
To alleviate this, we deliberately filter out machine-generated annotations and collect only samples with manual annotations (e.g., short explanations or forgery masks), ensuring a low-hallucination cold start.

\noindent \textbf{SFT data construction.}
Specifically, the samples are collected from SynthScars~\cite{kang2025legion}, AbHuman~\cite{fang2024humanrefiner}, SynthArtifact~\cite{cao2024synartifact}, X-AIGD~\cite{xiao2026unveiling} and the MiPO set of HydraFake~\cite{tan2025veritas}.
These samples encompass different types of annotation formats,
including detailed CoT trajectories (i.e., HydraFake), brief explanations (i.e., SynthScars) and artifacts descriptions with forged region bounding boxes (i.e., Abhuman, SynthArtifact and X-AIGD).
Real images are also collected from the aforementioned datasets and further supplemented with images from Flickr~\cite{young2014image} and VisualGenome~\cite{krishna2017visual}.
To unify these heterogeneous annotations into our pattern-aware reasoning format,
we employ Gemini-2.5-Pro~\cite{comanici2025gemini} to rewrite the original annotations.
We strictly prompt the model to perform logical restructuring without altering or fabricating the original evidence.
To further ensure data quality after rewriting, we take GPT-5.1~\cite{singh2025openai} to conduct quality filtering from two aspects:
(1) verifying whether the converted trajectory remains consistent with the original human annotation,
and (2) filtering out samples that contain a large number of weak descriptions (e.g., ``the image appears too perfect'').
Ultimately, we retain only the samples with precise reasoning and concrete descriptions, yielding approximately $10$K high-quality samples.

\noindent \textbf{Training objective.}
Suppose the cold-start dataset is denoted as $\mathcal{D}_1=\{(\bm{q}, \bm{s})_i\}_{i=1}^{N_1}$, where $\bm{s}$ is the target output sequence including pattern-aware reasoning and final answer. $\bm{q}$ denotes input image and user query.
Given $\mathcal{D}_1$, we perform supervised fine-tuning on the base MLLM $\pi_{\theta_0}$.
The training objective maximizes the likelihood of generating $\bm{s}$ given input $\bm{q}$:
\begin{equation}
    \mathcal{L}_{\mathrm{cold}}
    = -\mathbb{E}_{(\bm{q},\bm{s})\thicksim\mathcal{D}_1}
      \sum_{t=1}^{T}\log\pi_\theta(\bm{s}_t\mid\bm{q},\bm{s}_{<t}),
    \label{eq:cold}
\end{equation}
where $\pi_\theta$ denotes the token distribution of the current model and $\bm{s}_{<t}$ denotes the preceding tokens.
The resulting model $\pi_{\theta_{\mathrm{cold}}}$ carries a clean reasoning prior, serving as a reliable foundation of the subsequent learning stage.

\subsection{Perception-Oriented Learning}
\label{sec:method_perc}

As illustrated in Table~\ref{tab:perception_gap}, detection-targeted training still suffers from perception bottlenecks in capturing fine-grained visual details, semantic anomalies and pixel-level discrepancies.
To bridge this gap, we introduce Perception-oriented Learning, which replaces the open-ended artifact description objective with \textit{verifiable} perception rewards.
As shown in Figure~\ref{fig:method}, we consider three types of tasks: (1) fine-grained visual perception, (2) semantic anomaly perception and (3) pixel discrepancy perception.

\noindent \textbf{Fine-grained visual perception.}
Target visual clues may only occur in a spatially restricted region.
To improve the model's ability to capture small visual objects, 
we use localized visual questions that ask about the attribute, content or spatial relation of a target region.
Specifically, we randomly sample data from ZwZ~\cite{wei2026zooming} and Vero~\cite{sarch2026vero}, which contain detailed $\langle \text{image, query, answer} \rangle$ triplets for generic visual perception.
Given the reference answer $\bm{a}_f$ and the prediction $\hat{\bm{a}}_f$, the reward is calculated as:
\begin{equation}
    R_{\mathrm{fine}}=\mathbb{I}\bigl[\mathcal{N}(\hat{\bm{a}}_f)\simeq\mathcal{N}(\bm{a}_f)\bigr],
    \label{eq:rdetail}
\end{equation}
where $\mathcal{N}(\cdot)$ parses and normalizes the answer.
$\mathbb{I}[\cdot]$ denotes the indicator function, which outputs $1$ if the predicted answer $\hat{\bm{a}_f}$ is equivalent to the ground truth $\bm{a}_f$ and $0$ otherwise.
For $\simeq$, we first apply deterministic matching.
However, strict exact matching may yield false negatives when the model’s perception is correct but the output format deviates slightly from the ground truth (e.g., synonymous phrasing).
To avoid introducing noisy perception signals in such cases, we adopt LLM-as-a-Judge as a fallback for more robust reward calculation.
As a result, accurately answering these questions requires the model to attend to small designated regions rather than relying on global image semantics.

\noindent \textbf{Semantic anomaly perception.}
AI-generated images may violate anatomical and structural constraints, while current models fall short on perceiving these anomalies.
To mitigate this, we introduce both region-level anatomical judgment and image-level anomaly recognition.

\noindent \textbf{(1)} For region-level judgment, the training data is sampled from AbHuman~\cite{fang2024humanrefiner}.
We reorganize them into $\langle \text{image, box, query, answer} \rangle$ tuples, where the box denotes the localized body region, the query asks the model to determine the enclosed body part is normal, and the answer provides ground truth label (e.g., \textit{normal hand} or \textit{abnormal leg}).

\noindent \textbf{(2)} For image-level recognition, the training data is sampled from MagicData~\cite{wang2025magicmirror}.
The data is organized into $\langle \text{image, query, L1-answer, L2-answer} \rangle$ tuples, where the query asks the model to determine whether the image is normal (i.e., L1 level) and further predicts the corresponding anomaly categories such as element attributes, object interactions and human anatomy (i.e., L2 level). The answers provide ground truth labels for both levels.
Given the prediction $\hat{\bm{a}}_s$ and ground-truth structured labels $\bm{a}_s$,
the reward can be uniformly defined as:
\begin{equation}
    R_{\mathrm{sem}}=\mathbb{I}\bigl[\mathcal{N}(\hat{\bm{a}}_s)\simeq\mathcal{N}(\bm{a}_s)\bigr].
    \label{eq:rsemantic}
\end{equation}
By explicitly optimizing the reward, the model is incentivized to build a grounded understanding of physical and anatomical plausibility, enabling it to precisely identify genuine anomalies without misclassifying plausible visual content.

\noindent \textbf{Pixel discrepancy perception.}
Even when high-level content is coherent, the generated images may exhibit pixel-level inconsistency.
However, pixel-level perception is inherently ambiguous and eludes precise definition, as it often manifests as low-level signals such as noise patterns and frequency statistics, which are challenging to describe through language reasoning and hard to generalize across scenarios.
Rather than explicitly characterizing these signals, we aim to equip the model with the ability to perceive subtle discrepancies, as illustrated in Figure~\ref{fig:perception_example} (b).
This ability provides a foundation for detecting pixel-level anomalies.

\noindent To this end, we leverage images from Next-IMDL~\cite{li2025next} to construct the training data.
Specifically, the images consist of authentic and locally inpainted counterparts.
The model is required to identify regions whose pixel distributions are inconsistent with their surroundings and to respond with ``None'' when the whole image remains consistent.
Given the ground-truth box $\bm{b}$ and prediction $\hat{\bm{b}}$, the reward for localization are defined as the Intersection over Union (IoU):
\begin{equation}
    R_{\mathrm{loc}}(\hat{\bm{b}}, \bm{b})
    =\frac{|\hat{\bm{b}} \cap \bm{b}|}{|\hat{\bm{b}} \cup \bm{b}|} +\phi(\hat{\bm{b}},\bm{b}),
    \label{eq:bbox_reward}
\end{equation}
where $\phi(\cdot)$ is an area penalty that prevents the model from hacking overly large bounding boxes:
\begin{equation}
\phi(\hat{\bm{b}},\bm{b})=
\begin{cases}
0, & \rho(\hat{\bm{b}},\bm{b})\leq\gamma,\\
-\delta\left(\left\lfloor\log_2\dfrac{\rho(\hat{\bm{b}},\bm{b})}{\gamma}\right\rfloor+1\right),
& \rho(\hat{\bm{b}},\bm{b})>\gamma,
\end{cases}
\label{eq:area_penalty}
\end{equation}
where $\rho(\hat b,b)=\frac{|\hat{\bm{b}}|}{|\bm{b}|}$, and $\gamma=2.0$ and $\delta=0.3$ control the tolerance and penalty step, respectively.
To encourage joint reasoning about local pixel discrepancies and the overall authenticity, we augment the task by requiring the model to predict an overall image authenticity label. The reward is defined as:
\begin{equation}
    R_{\mathrm{label}}=\mathbb{I}\bigl[\mathcal{N}(\hat{\bm{a}}_l)\simeq\mathcal{N}(\bm{a}_l)\bigr].
\end{equation}
The complete reward for pixel perception task is calculated as:
\begin{equation}
    R_{\mathrm{pix}}=R_{\mathrm{loc}}+R_{\mathrm{label}}.
    \label{eq:rpixel}
\end{equation}
For authentic images, the ground-truth box is empty and $R_{\mathrm{loc}}$ is activated only for a valid fake prediction.

\noindent \textbf{Overall objective.}
Our method is suitable for most R1-style algorithms, and we adopt GSPO~\cite{zheng2025group} for training.
Suppose the training data is $\mathcal{D}_2 = \{(\bm{q}, \bm{a})_i \}_{i=1}^{N_2}$.
For each perception query $\bm{q}$, the model samples a group of responses $\{\hat{\bm{a}}_1, \hat{\bm{a}}_2, ..., \hat{\bm{a}}_G\}$, and the corresponding task verifier assigns rewards according to Eqs.~\eqref{eq:rdetail}, \eqref{eq:rsemantic}, and \eqref{eq:rpixel}.
A shared soft over-length reward $R_{\mathrm{len}}$ applies across all sub-tasks to discourage unnecessarily verbose reasoning without limiting legitimate analysis.
For a response of length $T$,
\begin{equation}
    R_{\mathrm{len}}=\min\!\left(
    -\frac{T-(L_{\max}-L_{\mathrm{cache}})}{L_{\mathrm{cache}}},\,0
    \right),
    \label{eq:rlen}
\end{equation}
where $L_{\max}$ is the maximum response length and $L_{\mathrm{cache}}$ is the soft cache window.
The unified reward for perception learning is:
\begin{equation}
    R = R_{k} + \lambda_{\mathrm{len}}\,R_{\mathrm{len}},
    \qquad k\in\{\mathrm{fine, sem, pix}\}.
    \label{eq:rtotal}
\end{equation}
Suppose the cold-started model $\pi_{\theta_{\text{cold}}}$ is adopted as reference policy,
the training objective is formulated as:
\begin{equation}
\begin{aligned}
\mathcal{L}_{\mathrm{PoRL}}
={}&-\mathbb{E}_{\substack{
    (\bm q,\bm a,k)\sim\mathcal D_2,\\
    \{\hat{\bm a}_i\}_{i=1}^{G}
    \sim\pi_{\theta_{\mathrm{old}}}(\cdot\mid\bm q)
}}
\left[
    \frac{1}{G}\sum_{i=1}^{G}\ell_i(\theta)
\right],
\\
\ell_i(\theta)
={}&\min\Big(
    s_i(\theta)A_i,\,
    \operatorname{clip}\!\left(
        s_i(\theta),1-\epsilon,1+\epsilon
    \right)A_i
\Big),
\end{aligned}
\label{eq:porl}
\end{equation}
where 
\begin{equation}
\begin{aligned}
    &s_i(\theta)=
\left(
\frac{
    \pi_\theta(\hat{\bm a}_i\mid\bm q,\hat{\bm a}_{i,<t})
}{
    \pi_{\theta_{\mathrm{old}}}
    (\hat{\bm a}_i\mid\bm q,\hat{\bm a}_{i,<t})
}
\right)^{\tfrac{1}{|\hat{\bm a}_i|}},\\
    &A_i=\frac{R_i-\mathrm{mean}(\{R_1,\ldots,R_G\})}{\mathrm{std}(\{R_1,\ldots,R_G\})}.
\end{aligned}
\end{equation}

\subsection{Value-Aware On-Policy Distillation}
\label{sec:vaopd}

To seamlessly integrate the perception capacities with authenticity reasoning, we take the advantages of On-Policy Distillation (OPD) to internalize the perception-aware reasoning.
However, we find that vanilla OPD inadequately models the varying value of the distillation signals.
We therefore propose Value-aware On-Policy Distillation (VaOPD) to close the gap.

\subsubsection{Preliminary: Privileged On-Policy Distillation}
For each training pair $(x_i,c_i)$, the student samples a reasoning trajectory under the standard query $\bm{q}_s$:
\begin{equation}
    \hat{\bm{a}}_i\sim p_{\theta}(\cdot\mid \bm{x}_i,\bm{q}_s).
    \label{eq:on_policy_rollout}
\end{equation}
Conditioned on each visited prefix $\hat{\bm{a}}_{i,<t}$, the student distribution and the privileged teacher distribution are denoted as:
\begin{equation}
\begin{aligned}
    \bm{S}_{i,t}&=p_{\theta}(\cdot\mid \bm{x}_i,\bm{q}_s,\hat{\bm{a}}_{i,<t}),\\
    \bm{T}_{i,t}&=p_{\bar\theta}(\cdot\mid \bm{x}_i,\bm{q}_t,\hat{\bm{a}}_{i,<t}),
    \label{eq:student_teacher_dist}
\end{aligned}
\end{equation}
where $\bm{q}_t$ augments $\bm{q}_s$ with training-only privileged information.
Unlike offline rationale distillation, the teacher does not generate a separate response.
It provides dense token distributions on the trajectories of the student.
Vanilla OPD treats these states uniformly and minimizes the Jensen--Shannon divergence (JSD) between $\bm{T}_{i,t}$ and $\bm{S}_{i,t}$.

\subsubsection{Empirical Motivation}
\label{sec:vaopd_motivation}

\begin{figure}[t]
    \centering
    \includegraphics[width=1.0\linewidth]{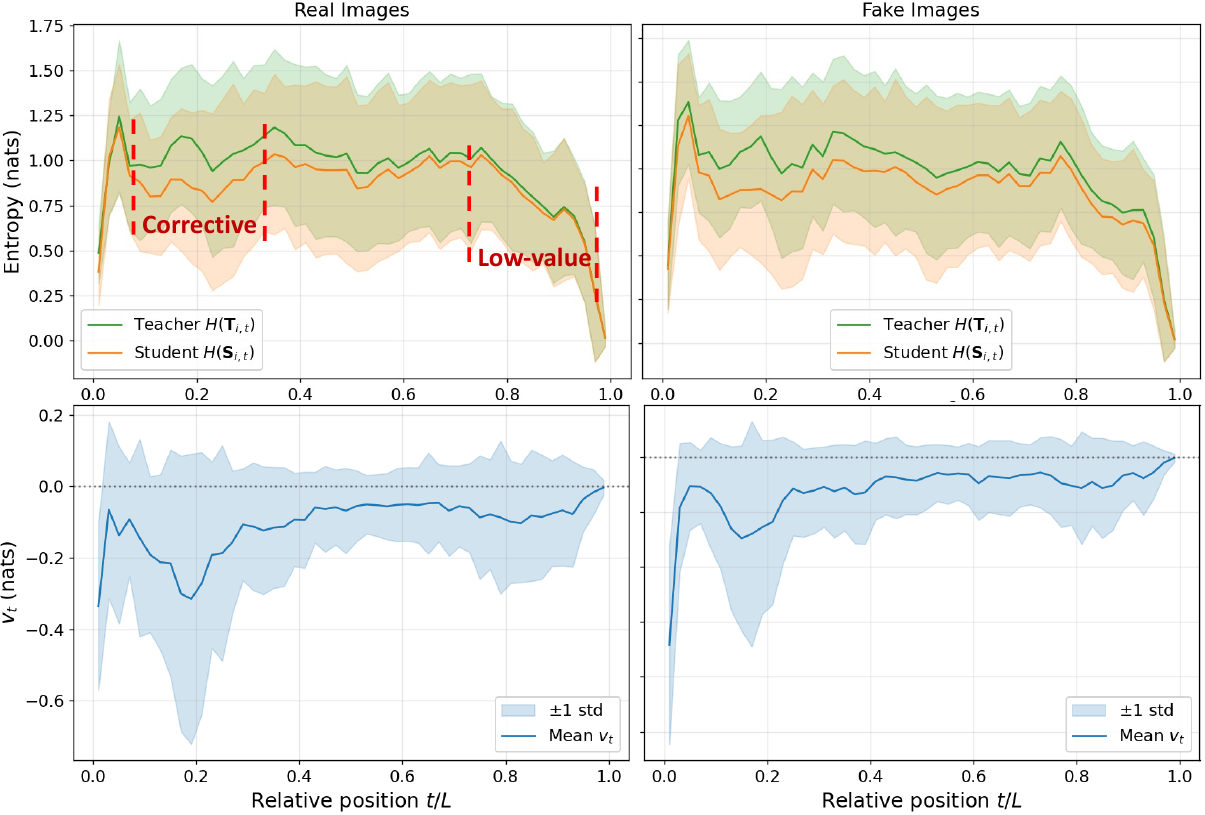}
    \caption{\textbf{Empirical motivation for VaOPD.}
    Token-wise teacher and student entropy (top) and teacher advantage $v_t$ (bottom) vary substantially along the reasoning process.
    Shaded regions denote one standard deviation.
    Corrective signals concentrate at positions with higher teacher entropy and negative $v_t$, whereas late tokens are often low-value.}
    \label{fig:vaopd_find}
\end{figure}

\begin{figure*}[t]
    \centering
    \includegraphics[width=0.99\linewidth]{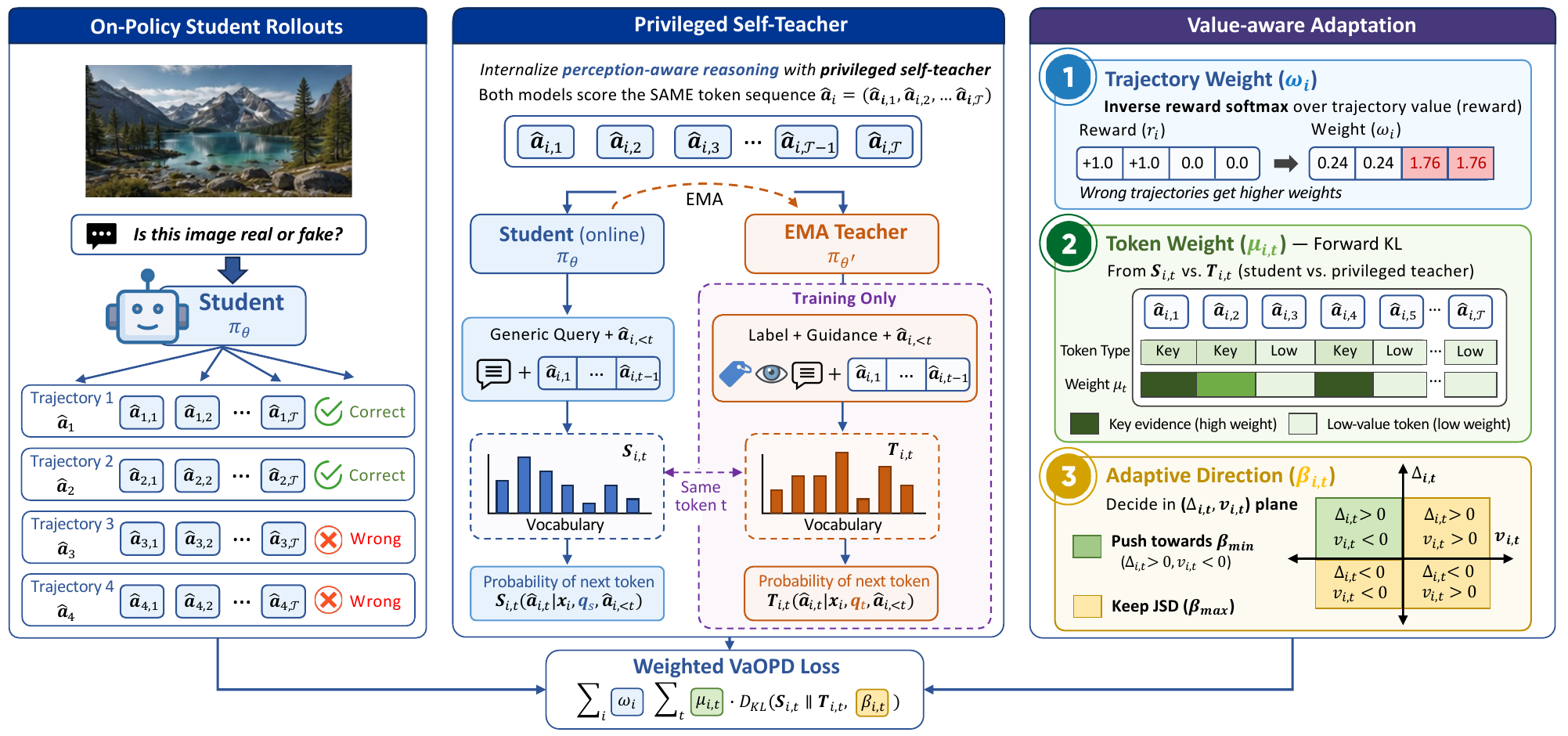}
    \vspace{-5pt}
     \caption{\textbf{Overview of Value-aware On-Policy Distillation (VaOPD).}
     \textbf{Left}: The student samples multiple on-policy reasoning trajectories under the standard authenticity query.
     \textbf{Middle}: An EMA self-teacher, additionally conditioned on training-only labels and perceptual guidance, scores the same student-generated token prefixes and provides dense privileged supervision without generating separate responses.
     \textbf{Right}: VaOPD models the varying value of these signals at three levels: trajectory weighting prioritizes incorrect rollouts, token weighting emphasizes positions with larger teacher--student discrepancies, and adaptive direction strengthens knowledge transfer at corrective tokens.}
	\label{fig:vaopd}
    \vspace{-0.4cm}
\end{figure*}

Standard OPD uniformly averages the distillation loss over response tokens, implicitly assuming that privileged information contributes equally throughout a reasoning trajectory.
We examine this by probing the teacher-and-student gap on the same student rollout.
Specifically, we evaluate the entropy difference $\Delta H_t$ and the teacher's advantage $v_t$ at each token position:
\begin{equation}
\begin{aligned}
    \Delta H_{i,t}
    &= H(\bm{T}_{i,t})-H(\bm{S}_{i,t}), \\
    v_{i,t}
    &= \log \bm{T}_{i,t}\!\left(\hat{a}_{i,t}\right)
      -\log \bm{S}_{i,t}\!\left(\hat{a}_{i,t}\right).
\end{aligned}
\label{eq:empirical_token_signals}
\end{equation}
As shown in Figure~\ref{fig:vaopd_find}, distillation value is unevenly distributed along a trajectory, i.e., late-stage tokens can be strongly constrained by an erroneous prefix and receive little new information from the teacher, as $\Delta H_{i,t}$ and $v_{i,t}$ both remain close to $0$.
Moreover, the privileged teacher exhibits higher predictive entropy at some positions, exposing visual evidence and reasoning alternatives \textit{not} covered by the student.
At such informative positions, the teacher tends to \textit{reject} the sampled token and redirect the trajectory, showing strong corrective signals.
These observations suggest three types of tokens.
\textit{(1) Low-value tokens} contain little privileged information and should be downweighted.
\textit{(2) Supportive tokens}, where the teacher support the current choice.
\textit{(3) Corrective tokens}, where the teacher rejects the student's current choice while exposing broader alternatives, which provides high distillation values for redirecting the reasoning.
Building on these observations, we introduce VaOPD to model distillation value from two complementary dimensions, i.e., reweighting and directional adaptation, as shown in Figure~\ref{fig:vaopd}.

\noindent \textbf{Trajectory and Token Reweighting.}
Correct rollouts already reflect successful perception, whereas erroneous ones expose perception-reasoning gaps that need correction.
Let $r_i=\mathbb{I}[\hat{\bm{a}}_i=\bm{a}_i]$ denotes the correctness of final verdict.
We prioritize wrong rollouts via an inversed softmax.
The trajectory-level weights $\omega_i$ are written as:
\begin{equation}
    \omega_i
    = B\cdot\frac{\exp(-r_i/\tau_r)}
                 {\sum_{j=1}^{B}\exp(-r_j/\tau_r)},
    \label{eq:traj}
\end{equation}
where $B$ is the number of rollouts in the batch and the leading factor $B$ keeps the mean trajectory weight at one.
Unlike hard filtering, we emphasizes incorrect responses while preserving correctly solved samples to maintain a balanced signal.

\noindent Diving into each token, we measure the privileged information introduced at each position through the teacher-to-student KL divergence.
Specifically, a large divergence indicates that the privileged teacher assigns substantial probability mass to continuations underrepresented by the student, thereby exposing a richer set of potential reasoning paths.
The token-level weights $\mu_{i,t}$ are calculated as:
\begin{equation}
    \mu_{i,t}
    =\mathcal{|T|}\cdot\frac{D_{\mathrm{KL}}(\bm{T}_{i,t}\|\bm{S}_{i,t})}
    {\sum_{(j,s)\in\mathcal{T}}\bigl[D_{\mathrm{KL}}(\bm{T}_{j,s}\|\bm{S}_{j,s})\bigr]}.
    \label{eq:token_weight}
\end{equation}

\noindent \textbf{Adaptive Distillation Direction.}
Token informativeness alone does not specify how the information should be transferred.
We therefore take the entropy difference $\Delta H_{i,t}$ and teacher advantage $v_{i,t}$ in Eq.~\eqref{eq:empirical_token_signals} to identify corrective tokens.
Specifically, we define the corrective value as:
\begin{equation}
    h_{i,t}
    = \Big[\frac{H(\bm{T}_{i,t})-H(\bm{S}_{i,t})}{\log|\mathcal{V}|}\Big]_+ \cdot[-v_{i,t}]_+,
    \label{eq:corrective_value}
\end{equation}
where $|\mathcal{V}|$ is the vocabulary size, and $[x]_+=\max(x,0)$ denotes the positive-part operator.
$h_{i,t}$ becomes positive only when the teacher exposes additional predictive modes while rejecting the student's sampled token.
The value is then normalized within the batch:
\begin{equation}
\begin{aligned}
    \bar h
    &=\frac{1}{|\mathcal{T}|}
      \sum_{(j,s)\in\mathcal{T}}h_{j,s},\\
    \rho_{i,t}
    &=\frac{h_{i,t}}{h_{i,t}+\bar h}.
\end{aligned}
\label{eq:normalized_corrective_value}
\end{equation}
Following GKD~\cite{agarwal2024onpolicy}, the distillation objective can be formulated as:
\begin{equation}
\begin{aligned}
    \bm{M}_{i,t}
    &=(1-\beta_{i,t})\bm{S}_{i,t}
      +\beta_{i,t}\bm{T}_{i,t},\\
    \mathcal{D}_{i,t}
    &=\beta_{i,t}D_{\mathrm{KL}}(\bm{T}_{i,t}\|\bm{M}_{i,t})\\
    &\quad +(1-\beta_{i,t})
    D_{\mathrm{KL}}(\bm{S}_{i,t}\|\bm{M}_{i,t}),
\end{aligned}
\label{eq:adaptive_jsd}
\end{equation}
where $\beta_{i,t}$ controls the distillation direction of each token.
For most existing methods, a fixed direction is adopted (e.g., $\beta_{i,t}=0.5$ or $\beta_{i,t}$=1.0).
In our case, we take corrective indicator $\rho_{i,t}$ to achieve adaptive directions, which can be formulated as:
\begin{equation}
    \beta_{i,t}
    =\beta_{\max}-(\beta_{\max}-\beta_{\min})\rho_{i,t}.
    \label{eq:adaptive_beta}
\end{equation}
A stronger corrective signal pushes $\beta_{i,t}$ toward $\beta_{\min}$, shifting the objective toward forward KL to cover alternative reasoning modes exposed by the privileged teacher.

\noindent \textbf{Overall optimization.}
Combining these together, the overall VaOPD objective is formulated as:
\begin{equation}
    \mathcal{L}_{\mathrm{VaOPD}}
    =\frac{1}{\sum_{i=1}^{B}\omega_i}
    \sum_{i=1}^{B}\omega_i
    \left[
    \frac{1}{T_i}\sum_{t=1}^{T_i}
    \mu_{i,t}\mathcal{D}_{i,t}
    \right].
    \label{eq:vaopd_objective}
\end{equation}
To provide stable supervision without relying on an external teacher, we update the teacher after each student optimization step using an exponential moving average (EMA):
\begin{equation}
    \bar\theta\leftarrow\eta\bar\theta+(1-\eta)\theta.
    \label{eq:ema}
\end{equation}

\section{Experiments}
\subsection{Experimental Setup}

\noindent \textbf{Evaluation benchmarks.}
To rigorously assess generalization, we organize the detection benchmarks into three progressively challenging regimes.
\textbf{(1) Standard benchmarks} include AIGI-Now~\cite{chen2025task}, Chameleon~\cite{yan2025sanity}, GenBuster++~\cite{wen2025busterx++}, EvalGen~\cite{chen2025dual}, HydraFake (cross-forgery set)~\cite{tan2025veritas} and HiRes~\cite{mu2025no}, covering diverse generators and commonly studied forgery settings.
\textbf{(2) In-the-wild benchmarks} comprise RealChain~\cite{liu2025beyond}, CommunityAI~\cite{li2026artificial}, SocialRF~\cite{li2026artificial} and WildRF~\cite{cavia2024real}, which better reflect the diverse content and uncontrolled processing encountered in real-world media.
\textbf{(3) Emerging generators and scenarios} include GPT-Image-2 and the text-rich forgery benchmark TFR~\cite{qu2026textshield}.
For GPT-Image-2, we collect $3$K samples from GPT-Image-2 Wild~\cite{zewde2026gpt} and online platforms.
For TFR, we focus on the whole-generated subset.
This regime evaluates not only generalization to newly developed generators and semantic domains, but also the model's ability to continually acquire new capabilities.

\begin{table*}[t]
    \scriptsize
    \centering
    \caption{Performance comparison (Acc.) on Standard and Emerging OOD AIGI detection benchmarks.
    HR. and ANow. denote HiRes~\cite{mu2025no} and AIGI-Now~\cite{chen2025task}, respectively.
    \textit{900} and \textit{2000} denote the resolution of \textit{0-900} and \textit{1500-2000}, refer to HiRes~\cite{mu2025no}.
    $\dagger$ denotes we incorporate samples from new scenarios for training to evaluate model's capability to adapt toward emerging generators and scenarios.
    The best results are \textbf{bolded} and the second best are \underline{underlined}.}
    \vspace{-3pt}
    \label{tab:aigc_detection}
    \renewcommand{\arraystretch}{1.0}
    \setlength{\tabcolsep}{3.2pt}
    \resizebox{\textwidth}{!}{
        \begin{tabular}{
            p{56pt}<{\raggedright}                         
            p{34pt}<{\centering}                           
            p{30pt}<{\centering}                           
            p{28pt}<{\centering}                           
            p{30pt}<{\centering}                           
            p{30pt}<{\centering}                           
            p{31pt}<{\centering}                           
            p{32pt}<{\centering}                           
            p{32pt}<{\centering}                           
            p{20pt}<{\centering}
            p{28pt}<{\centering}                           
            p{32pt}<{\centering}                           
            p{20pt}<{\centering}                           
        }
            \toprule
            \multirow{2}{*}{\textbf{Method}}
            & \multicolumn{8}{c}{\textbf{Standard OOD}}
            & \multirow{2}{28pt}{\hspace{-6pt}\centering\textbf{Avg.}}
            & \multicolumn{2}{c}{\textbf{Emerging OOD}}
            & \multirow{2}{28pt}{\hspace{-6pt}\centering\textbf{Avg.}} \\
            \cmidrule(lr){2-9}
            \cmidrule(lr){11-12}
            & Chameleon
            & GenB.++
            & EvalGen
            & HydraF.
            & HR.-\textit{900}
            & HR.-\textit{2000}
            & ANow.-\textit{Pix}
            & \hspace{-3pt}ANow.-\textit{Sem}
            & 
            & TFR
            & \hspace{-3pt}GPT-Img.2
            & \\
            
            \shline
            \rowcolor{lightblue}\multicolumn{2}{c}{\hspace{-33pt}\textit{\textbf{Vision-only Detectors}}} &&&&&&&&&&&\\
            AIDE~\cite{yan2025sanity} & 65.7 & 55.9 & 20.0 & 62.5 & 59.3 & 63.0 & 77.1 & 57.3 & 57.6 & 52.4 & 49.3 & 50.8 \\
            FatFormer~\cite{liu2024forgery} & 58.0 & 50.0 & 27.0 & 68.4 & 50.4 & 50.7 & 53.2 & 50.0 & 51.0 & 50.0 & 51.7 & 50.9 \\
            DRCT~\cite{chen2024drct} & 70.6 & 48.8 & 50.6 & 78.1 & 80.3 & 76.8 & 70.4 & 59.3 & 66.9 & 47.1 & 41.8 & 44.5 \\
            D3QE~\cite{zhang2025d3qe} & 56.1 & 49.8 & 27.0 & 61.5 & 51.2 & 52.7 & 52.2 & 48.2 & 49.8 & 51.2 & 48.3 & 49.8 \\
            SPAI~\cite{karageorgiou2025any} & 59.6 & 71.4 & 90.3 & 69.7 & 76.7 & 70.7 & 82.2 & 62.8 & 72.9 & 36.8 & 45.6 & 41.2 \\
            PGC~\cite{zhou2026pgc} & 61.8 & 48.1 & 57.6 & 77.6 & 52.5 & 66.7 & 64.6 & 58.9 & 61.0 & 48.4 & 60.1 & 54.3 \\
            AlignedForen~\cite{rajanaligned} & 69.8 & 63.4 & 52.6 & 62.1 & 88.7 & 80.9 & 65.6 & 49.9 & 66.6 & 49.8 & 49.0 & 49.4 \\
            B-Free~\cite{guillaro2025bias} & 78.7 & 52.0 & 90.8 & \underline{92.1} & 79.8 & 88.9 & 71.0 & 53.2 & 75.8 & \textbf{78.9} & 49.5 & 64.2 \\
            DDA~\cite{chen2025dual} & 84.4 & 52.5 & 97.2 & 88.7 & 89.4 & 87.4 & 83.4 & 55.8 & 79.9 & 64.8 & 53.0 & 58.9 \\
            Co-SPY~\cite{cheng2025co} & 71.1 & 56.9 & 92.4 & 74.1 & 76.0 & 59.5 & 80.8 & 69.5 & 72.5 & 50.4 & 56.4 & 53.4 \\
            Effort-AIGI~\cite{yan2025orthogonal} & 63.9 & 41.0 & 97.6 & 51.2 & 60.8 & 57.6 & 75.5 & 64.9 & 64.1 & 57.2 & 58.4 & 57.8 \\
            OmniAID~\cite{guo2025omniaid} & \underline{94.2} & 64.6 & 97.1 & \textbf{95.3} & \textbf{94.1} & 90.1 & 78.0 & 91.2 & 88.1 & 64.0 & 71.8 & 67.9 \\

            \shline
            \rowcolor{lightblue}\multicolumn{2}{c}{\textit{\textbf{MLLM-based Forgery Detectors}}} &&&&&&&&&&&\\
            BusterX++~\cite{wen2025busterx++} & 76.3 & 77.8 & \textbf{99.5} & 70.2 & 87.6 & 85.8 & 95.4 & 94.0 & 85.8 & 58.1 & \underline{79.4} & 68.8 \\
            DeepVRM~\cite{lin2026deep} & \textbf{95.1} & 49.5 & \underline{97.8} & 54.3 & 84.7 & 76.9 & 97.0 & 94.4 & 81.2 & 54.5 & 71.7 & 63.1 \\
            FakeShield~\cite{xu2025fakeshield} & 49.2 & 48.0 & 94.6 & 63.4 & 63.0 & 53.8 & 96.8 & 76.4 & 68.2 & 50.4 & 56.8 & 53.6 \\
            FakeVLM~\cite{wen2026spot} & 63.0 & 55.6 & \textbf{99.5} & 76.6 & 65.7 & 64.3 & 95.7 & 90.0 & 76.3 & 57.6 & 58.4 & 58.0 \\
            UniGenDet~\cite{zhang2026unigendet} & 73.4 & 63.1 & 70.6 & 59.6 & 81.1 & 78.9 & 84.4 & 86.5 & 74.7 & 58.1 & 58.7 & 58.4 \\
            VideoVeritas~\cite{tan2026videoveritas} & 79.9 & 85.6 & 97.5 & 80.1 & 89.8 & 92.4 & 91.1 & 91.9 & 88.5 & 59.9 & 62.6 & 61.3 \\

            \shline
            \rowcolor{lightblue}\multicolumn{2}{c}{\hspace{-6pt}\textit{\textbf{Base and Preliminary Models}}} &&&&&&&&&&&\\
            Qwen3-VL-8B~\cite{bai2025qwen3} & 77.3 & 61.2 & 80.8 & 62.3 & 74.6 & 71.1 & 73.6 & 90.8 & 74.0 & 61.4 & 64.4 & 62.9 \\
            \veritas~\cite{tan2025veritas} & 68.2 & 66.4 & 84.3 & 90.3 & 83.2 & 67.1 & 73.6 & 84.8 & 77.2 & 38.7 & 71.6 & 55.1 \\
            \veritaspp & 88.9 & \textbf{92.8} & \textbf{99.5} & 88.1 & \underline{92.8} & \textbf{95.7} & \textbf{97.8} & \textbf{96.7} & \textbf{94.0} & 67.0 & 73.6 & \underline{70.3} \\
            \rowcolor{Gray} \textbf{\veritaspp$^{\dagger}$} & 91.1 & \underline{90.5} & \textbf{99.5} & 85.6 & \textbf{94.1} & \underline{93.8} & \underline{97.5} & \underline{96.6} & \underline{93.6} & \underline{76.4} & \textbf{84.8} & \textbf{80.6} \\

            \bottomrule
        \end{tabular}
    }
    \vspace{-0.1cm}
\end{table*}

\noindent \textbf{Comparison methods.}
We conduct a comprehensive comparison with existing AIGI detectors, reproducing their results using the officially released checkpoints and inference pipelines.
\textbf{(1) Vision-only methods} cover three representative technical directions: \textbf{a)} Low-level artifact modeling, such as AIDE~\cite{yan2025sanity} and SPAI~\cite{karageorgiou2025any}.
\textbf{b)} Bias-free training, such as B-Free~\cite{guillaro2025bias} and DDA~\cite{chen2025dual}.
\textbf{c)} Semantic decoupling, such as Effort~\cite{yan2025orthogonal} and OmniAID~\cite{guo2025omniaid}.
If multiple official variants are available, we select the strongest released checkpoint, e.g., we use the DINOv3-trained model for OmniAID.
\textbf{(2) MLLM-based methods} include reasoning-oriented BusterX++~\cite{wen2025busterx++}, the unified model UniGenDet~\cite{zhang2026unigendet}, and our preliminary work \veritas~\cite{tan2025veritas}, all following the official inference guidelines.

\noindent \textbf{Implementation details.}
We implement \veritaspp\ with Qwen3-VL-8B~\cite{bai2025qwen3}.
Cold-start is trained for $2$ epochs and other stages are trained for $1$ epoch.
For cold-start, the model is trained on $10$K samples for $2$ epochs.
For Perception-oriented Learning, the model is trained on $20$K samples for $1$ epoch, sampling $G=4$ responses at a temperature of $1.0$, with $L_{\text{max}}=1536$ and $L_{\text{cache}}=896$.
For VaOPD, the model is trained on $11$K samples for $1$ epoch.
EMA decay is set to $\eta=0.99$, the trajectory reweighting temperature $\tau_r=0.5$, and the adaptive JSD range $[\beta_{\min},\beta_{\max}]=[0.1,0.5]$.
All three stages are trained with LoRA~\cite{hu2022lora} (rank=$64$, $\alpha=128$) and a batch size of $64$.

\begin{table}[t]
    \scriptsize
    \centering
    \caption{Performance comparison (Acc.) on In-the-Wild AIGI detection benchmarks.
    Comm.AI denotes CommunityAI~\cite{li2026artificial}.
    $\dagger$ denotes capability evolution toward emerging generators and scenarios.
    The best results are \textbf{bolded} and the second best are \underline{underlined}.}
    \vspace{-3pt}
    \label{tab:real_world_detection}
    \renewcommand{\arraystretch}{1.0}
    \setlength{\tabcolsep}{3pt}
    \scalebox{1.04}{
        \begin{tabular}{
            p{56pt}<{\raggedright}
            p{32pt}<{\centering}
            p{30pt}<{\centering}
            p{30pt}<{\centering}
            p{30pt}<{\centering}
            p{25pt}<{\centering}
        }
            \toprule
            \multirow{2}{*}{\textbf{Method}}
            & \multicolumn{4}{c}{\textbf{In-the-Wild}}
            & \multirow{2}{*}{\textbf{Avg.}} \\
            \cmidrule(lr){2-5}
            & RealChain
            & Comm.AI
            & SocialRF
            & WildRF
            & \\
            \shline
            \multicolumn{6}{l}{
                \cellcolor{lightblue}
                \hspace{-2pt}\textit{\textbf{Vision-only Detectors}}
            } \\

            AIDE~\cite{yan2025sanity}
            & 54.4 & 66.2 & 62.4 & 69.1 & 63.0 \\

            FatFormer~\cite{liu2024forgery}
            & 48.9 & 51.9 & 56.9 & 63.7 & 55.4 \\

            DRCT~\cite{chen2024drct}
            & 65.0 & 76.7 & 67.5 & 70.5 & 69.9 \\

            D3QE~\cite{zhang2025d3qe}
            & 49.2 & 51.0 & 50.7 & 53.7 & 51.2 \\

            PGC~\cite{zhou2026pgc}
            & 51.6 & 59.9 & 77.1 & 85.4 & 68.5 \\

            AlignedForen.~\cite{rajanaligned}
            & 41.6 & 68.3 & 52.5 & 54.3 & 54.2 \\

            B-Free~\cite{guillaro2025bias}
            & 67.6 & 81.5 & 85.2 & 90.7 & 81.3 \\

            DDA~\cite{chen2025dual}
            & 69.0 & 84.7 & 81.8 & 88.5 & 81.0 \\

            Co-SPY~\cite{cheng2025co}
            & 70.0 & 66.5 & 68.3 & 75.8 & 70.2 \\

            Effort-AIGI~\cite{yan2025orthogonal}
            & 55.0 & 91.3 & 66.3 & 61.4 & 68.5 \\

            OmniAID~\cite{guo2025omniaid}
            & 75.5 & \underline{92.2} & \textbf{94.6}
            & \underline{96.2} & 89.6 \\

            \shline
            \multicolumn{6}{l}{
                \cellcolor{lightblue}
                \hspace{-2pt}\textit{\textbf{MLLM-based Forgery Detectors}}
            } \\

            BusterX++~\cite{wen2025busterx++}
            & 77.0 & 78.9 & 85.0 & 93.3 & 83.6 \\

            DeepVRM~\cite{lin2026deep}
            & 80.2 & \textbf{97.2} & 83.1 & 89.4 & 87.5 \\

            FakeShield~\cite{xu2025fakeshield}
            & 60.4 & 53.4 & 73.0 & 78.6 & 66.4 \\

            FakeVLM~\cite{wen2026spot}
            & 72.2 & 64.0 & 63.3 & 67.9 & 66.9 \\

            UniGenDet~\cite{zhang2026unigendet}
            & 74.4 & 65.7 & 75.6 & 81.8 & 74.4 \\

            VideoVeritas~\cite{tan2026videoveritas}
            & 76.6 & 80.6 & 86.0 & 90.7 & 83.5 \\

            \shline
            \multicolumn{6}{l}{
                \cellcolor{lightblue}
                \hspace{-2pt}\textit{\textbf{Base and Preliminary Models}}
            } \\

            Qwen3-VL-8B~\cite{bai2025qwen3}
            & 65.4 & 78.5 & 90.9 & 93.0 & 82.0 \\
            
            \veritas~\cite{tan2025veritas}
            & 56.6 & 70.0 & 78.5 & 87.0 & 73.0 \\

            \veritaspp
            & \textbf{82.3} & 87.6 & \underline{93.7}
            & \textbf{96.4} & \underline{90.0} \\

            \rowcolor{Gray}
            \textbf{\veritaspp$^\dagger$}
            & \underline{81.8} & 91.5 & 92.2
            & 96.0 & \textbf{90.4} \\

            \bottomrule
        \end{tabular}
    }
    \vspace{-0.1cm}
\end{table}

\subsection{Main Results}

\noindent \textbf{Comparison with vision-only detectors.}
As shown in Tables~\ref{tab:aigc_detection} and~\ref{tab:real_world_detection}, \veritaspp$\,$ establishes a clear performance advantage over existing vision-only detectors (e.g., +$5.9\%$ over previous best on standard OOD).
\textbf{(1)} Low-level methods such as AIDE~\cite{yan2025sanity} and SPAI~\cite{karageorgiou2025any} fail to achieve satisfactory generalization across the benchmarks.
\textbf{(2)} Bias-free training instead brings clear improvements, e.g., DDA~\cite{chen2025dual} reaches $79.9\%$ average accuracy on the standard benchmarks, while dropping to $58.9\%$ on emerging scenarios.
\textbf{(3)} In comparison, stronger visual foundation models substantially improves generalization.
OmniAID~\cite{guo2025omniaid} which is built upon DINOv3, achieves competitive performance across multiple settings, reaching $89.6\%$ on in-the-wild data.
In contrast, \veritaspp$\,$ exhibits more balanced performance across diverse detection benchmarks.
Notably, on perception-intensive AIGI-Now, \veritaspp$\,$ surpasses the best vision-only detectors by $14.4\%$ and $5.5\%$ on pixel-level and semantic forgeries, respectively.
These results demonstrate its advantage over vision-only detectors in learning robust and transferable perceptual evidence.

\noindent \textbf{Comparison with MLLM-based detectors.}
Existing MLLM-based detectors show strong performance on certain datasets, but their overall results remain less balanced.
\textbf{(1)} DeepVRM~\cite{lin2026deep} which explicitly considers pixel-level perception, achieves impressive accuracy on Chameleon (i.e., $95.1\%$) and an average of $87.5\%$ on in-the-wild benchmarks, but the overall performance remains moderate compared to \veritaspp$\,$ (e.g., $-12.8\%$ on standard OOD).
This suggests the advantage of comprehensive perception training for AIGI detection.
\textbf{(2)} UniGenDet~\cite{zhang2026unigendet} which adopts a co-evolution strategy, does not exhibit clear advantages on detection task.
\textbf{(3)} More importantly, compared with our preliminary \veritas, it improves the results by $16.8\%$, $17.0\%$ and $15.2\%$ on the standard, in-the-wild and emerging benchmarks, respectively.
These consistent improvements demonstrate that extending pattern-aware reasoning with explicit perception learning and VaOPD substantially enhances general AIGI detection.

\noindent \textbf{Capability evolution toward emerging scenarios.}
We further investigate whether \veritaspp$\,$ can acquire emerging detection capabilities without sacrificing its existing generalization.
As shown in Tables~\ref{tab:aigc_detection} and~\ref{tab:real_world_detection},
we incorporate a small number of samples from the training sets of TFR and GPT-Image-2 during VaOPD stage (denoted as \veritaspp$^\dagger$).
Notably, \veritaspp$^\dagger$ improves the performance on emerging scenarios from $70.3\%$ to $80.6\%$, 
with the performance on the standard benchmarks only decreases slightly from $94.0\%$ to $93.6\%$, 
while the in-the-wild average is maintained and even improves from $90.0\%$ to $90.4\%$.
Specifically, the performance improve significantly on TFR and GPT-Image-2 (+$9.4\%$ and $11.2\%$, respectively).
These results demonstrate that the proposed VaOPD can effectively acquire capabilities for novel generators and target scenarios while preserving the model's existing detection ability.

\begin{figure*}[t]
    \scriptsize
    \centering
	\begin{minipage}{0.49\linewidth}
            \captionof{table}{Ablation of the three-stage training pipeline. Results are averaged within each evaluation regime (\%).}
            \vspace{-3pt}
    \label{tab:abl_training_stages}
    \renewcommand{\arraystretch}{1.06}
    \setlength{\tabcolsep}{3pt}
    \resizebox{\columnwidth}{!}{
        \begin{tabular}{
            p{80pt}<{\raggedright}
            p{21pt}<{\centering}p{21pt}<{\centering}
            *{4}{p{20pt}<{\centering}}
        }
            \multirow{2}{*}{\textbf{Training Setting}}
            & \multicolumn{2}{c}{\textbf{Standard OOD}}
            & \multicolumn{2}{c}{\textbf{In-the-Wild}}
            & \multicolumn{2}{c}{\textbf{Emerging OOD}} \\
            \cmidrule(lr){2-3}
            \cmidrule(lr){4-5}
            \cmidrule(lr){6-7}
            & Acc & F1 & Acc & F1 & Acc & F1 \\
            \shline
            Base Model                  & 74.0 & 67.2 & 82.0 & 79.3 & 62.9 & 67.7 \\
            $+$ SFT                     & 89.7 & 89.4 & 87.1 & 86.9 & 63.4 & 54.7 \\
            $+$ SFT $+$ VaOPD           & \underline{92.2} & \underline{91.7} & 88.5 & \underline{88.8} & \underline{74.8} & \underline{73.5} \\
            $+$ SFT $+$ PoRL            & 92.0 & 91.2 & \underline{88.6} & 88.5 & 66.7 & 53.6 \\
            \rowcolor{Gray}$+$ SFT $+$ PoRL $+$ VaOPD
                                        & \textbf{93.6} & \textbf{92.4} & \textbf{90.4} & \textbf{89.9} & \textbf{80.6} & \textbf{77.0} \\
        \end{tabular}
    }
	\end{minipage}
    \begin{minipage}{0.49\linewidth}
            \captionof{table}{Ablation of the three perception-oriented data types. Results are averaged within each evaluation regime (\%).}
            \vspace{-3pt}
    \label{tab:abl_perception_data}
    \renewcommand{\arraystretch}{1.06}
    \setlength{\tabcolsep}{3pt}
    \resizebox{\columnwidth}{!}{
        \begin{tabular}{
            p{56pt}<{\raggedright}
            *{6}{p{22pt}<{\centering}}
        }
            \multirow{2}{*}{\textbf{Training Setting}}
            & \multicolumn{2}{c}{\textbf{Standard OOD}}
            & \multicolumn{2}{c}{\textbf{In-the-Wild}}
            & \multicolumn{2}{c}{\textbf{Emerging OOD}} \\
            \cmidrule(lr){2-3}
            \cmidrule(lr){4-5}
            \cmidrule(lr){6-7}
            & Acc & F1 & Acc & F1 & Acc & F1 \\
            \shline
            \rowcolor{Gray}\textit{w/} PoRL       & \textbf{93.6} & \underline{92.4} & \textbf{90.4} & \textbf{89.9} & \textbf{80.6} & \textbf{77.0} \\
            \hspace{6pt}\textit{w/o} Fine         & \underline{93.0} & \textbf{92.6} & 89.6 & \underline{89.6} & 78.9 & 74.0 \\
            \hspace{6pt}\textit{w/o} Semantic     & 92.3 & 91.4 & \underline{90.1} & \textbf{89.9} & \underline{80.0} & \underline{76.4} \\
            \hspace{6pt}\textit{w/o} Pixel        & 92.6 & 92.1 & 89.7 & 89.5 & 79.6 & 75.6 \\
            \textit{w/o} PoRL                     & 92.2 & 91.7 & 88.5 & 88.8 & 74.8 & 73.5 \\
        \end{tabular}
    }
	\end{minipage}
    \vspace{-0.3cm}
\end{figure*}

\subsection{Ablation Studies}

\noindent \textbf{Ablations on different training stages.}
As shown in Table~\ref{tab:abl_training_stages}:
\textbf{(1)} SFT provides a strong cold start, improving the average accuracy on standard benchmarks from $74.0\%$ to $89.7\%$.
However, its gain on the more challenging emerging benchmarks is marginal (i.e., $+0.5\%$ Acc), indicating that detection-oriented supervision alone is insufficient for newly emerging generators and scenarios.
\textbf{(2)} Although PoRL is optimized with perception-oriented tasks rather than detection supervision, it \textit{seamlessly} improves the detection accuracy by $2.3\%$ and $3.3\%$ on the standard and emerging benchmarks, respectively.
This verifies that enhanced perception provides a fundamental basis for generalizable detection.
\textbf{(3)} More importantly, perception learning further \textit{amplifies} the benefit of VaOPD.
On the emerging benchmarks, applying VaOPD after PoRL brings gains of $13.9\%$ Acc and $23.4\%$ F1, compared with $11.4\%$ and $18.8\%$ without PoRL.
This suggests that PoRL equips the privileged self-teacher with stronger perceptual capabilities, thereby providing more effective guidance for VaOPD to internalize perception into detection.

\noindent \textbf{Effect of different perception tasks.}
As shown in Table~\ref{tab:abl_perception_data}, we examine the contribution of each perception task by removing one branch at a time from the full PoRL setting.
\textbf{(1)} Removing any of the three perception tasks consistently degrades performance across all evaluation regimes, confirming that fine-grained visual, semantic anomaly and pixel-level perception are all beneficial to detection.
\textbf{(2)} Fine-grained perception contributes significantly to the emerging benchmarks (i.e., $+1.7\%$ Acc. and $+3.0\%$ F1), suggesting that the ability to attend to small regions is fundamental to detecting subtle artifacts introduced by novel generators.
\textbf{(3)} Removing the semantic anomaly task severely hurts the standard OOD performance (i.e., $-1.3\%$ Acc), implying that structural and anatomical plausibility serves as a transferable cue across common generators.
\textbf{(4)} Removing the pixel-level task yields broad degradation on both in-the-wild and emerging benchmarks, indicating that low-level distributional discrepancies remain informative when high-level semantics appear visually coherent.
These observations validate the importance of multiple levels of perception, which comprehensively strengthen the perceptual foundation for AIGI detection.

\noindent \textbf{Effect of different strategies in VaOPD.}
As shown in Table~\ref{tab:abl_vaopd_components}, we dissect VaOPD into three components, i.e., trajectory reweighting (i.e., Traj.), token reweighting (i.e., Tok.), and adaptive distillation direction (i.e., Dir.) and examine their individual and combined effects.
\textbf{(1)} Each individual component already brings gains over vanilla OPD on the emerging benchmarks, with token reweighting contributing the largest single improvement (i.e., $+3.8\%$ F1).
\textbf{(2)} Combining trajectory and token reweighting yields a notable synergy (i.e., $+5.1\%$ F1 on emerging over vanilla OPD), as prioritizing erroneous rollouts and informative tokens jointly focuses the supervision on the most learning-demanding positions.
\textbf{(3)} The full combination of all three components achieves further improvements across all regimes, demonstrating that adaptive distillation direction further complements reweighting by steering the objective toward alternative reasoning modes exposed by the privileged teacher.
These results confirm that all three components are essential and complementary in modeling the varying distillation value.

\begin{figure}[t]
    \centering
    \includegraphics[width=0.96\linewidth]{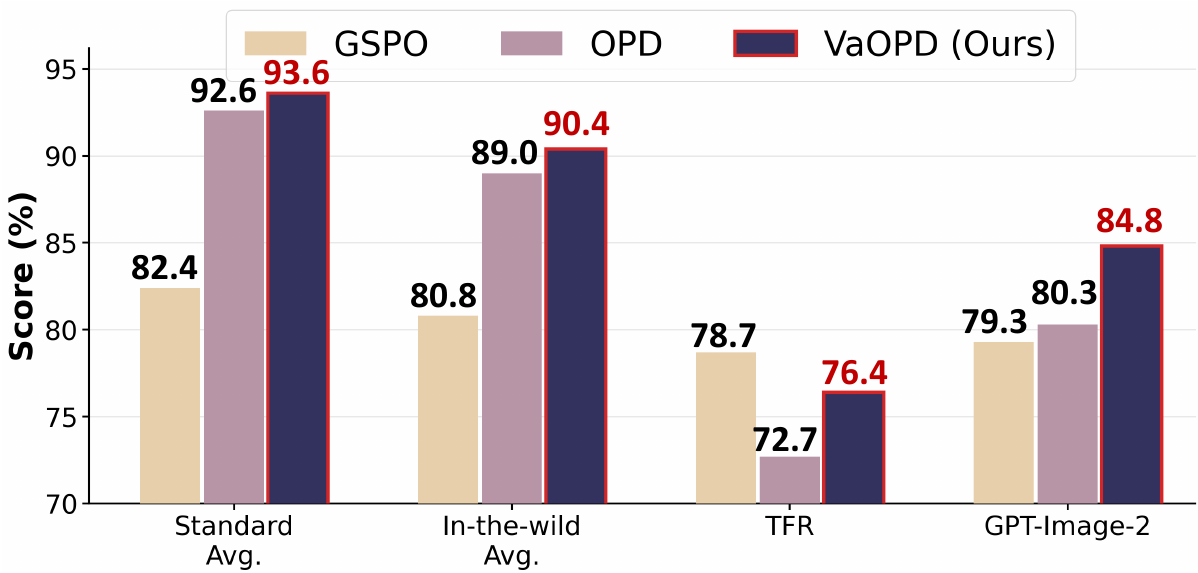}
    \vspace{-3pt}
     \caption{\textbf{Comparison of GSPO, vanilla OPD and VaOPD.}
     Standard and in-the-wild results are averaged over their corresponding benchmarks.
     VaOPD better balances existing generalization and capability acquisition on emerging scenarios.}
	\label{fig:improve}
    \vspace{-0.3cm}
\end{figure}

\noindent \textbf{Comparisons of RLVR, OPD and VaOPD.}
We further compare three learning paradigms built upon the same model, including RLVR (i.e., GSPO-style method), vanilla OPD and the proposed VaOPD.
As shown in Table~\ref{tab:abl_vaopd_components},
\textbf{(1)} RLVR achieves promising gains on the emerging benchmarks but incurs \textit{substantial forgetting} on the standard and in-the-wild benchmarks, preventing effective and sustainable capability evolution.
\textbf{(2)} Vanilla OPD steadily improves all performance, while the gains on emerging scenarios are limited.
\textbf{(3)} VaOPD consistently outperforms the other two paradigms across all three regimes, with significant gains on the emerging benchmarks (i.e., $+4.1\%$ Acc. and $+5.6\%$ F1 over vanilla OPD).
Figure~\ref{fig:improve} further shows the advantages of VaOPD on learning emerging capabilities.
This comparison demonstrates that explicitly modeling distillation value is \textit{critical} for transferring perception-aware reasoning into the model, especially when generalizing to \textit{unseen} generators and scenarios.

\begin{table}[t]
    \scriptsize
    \centering
    \caption{Ablations on VaOPD. ``Traj.'', ``Tok.'', and ``Dir.'' denote trajectory reweighting, token reweighting and adaptive direction. ``Base'' is the ``SFT$+$PoRL'' trained model.}
    \vspace{-3pt}
    \label{tab:abl_vaopd_components}
    \renewcommand{\arraystretch}{1.04}
    \setlength{\tabcolsep}{2.2pt}
    \resizebox{\columnwidth}{!}{
        \begin{tabular}{
            p{22pt}<{\raggedright}
            *{3}{p{16pt}<{\centering}}
            *{6}{p{18pt}<{\centering}}
        }
            \multirow{2}{*}{\hspace{-1pt}\textbf{Method}}
            & \multirow{2}{*}{\textbf{Traj.}}
            & \multirow{2}{*}{\textbf{Tok.}}
            & \multirow{2}{*}{\textbf{Dir.}}
            & \multicolumn{2}{c}{\textbf{Standard}}
            & \multicolumn{2}{c}{\textbf{In-the-Wild}}
            & \multicolumn{2}{c}{\textbf{Emerging}} \\
            \cmidrule(lr){5-6}
            \cmidrule(lr){7-8}
            \cmidrule(lr){9-10}
            & & & & Acc & F1 & Acc & F1 & Acc & F1 \\
            \shline
            \rule{0pt}{7pt}Base & -- & -- & -- & 92.0 & 91.2 & 88.6 & 88.5 & 66.7 & 53.6 \\
            GSPO & -- & -- & -- & 82.4 & 82.3 & 80.8 & 83.5 & 79.0 & \textbf{80.1} \\
            OPD  & \ding{55} & \ding{55} & \ding{55} & 92.6 & 91.8 & 89.0 & 89.1 & 76.5 & 71.4 \\
            \shline
            \rule{0pt}{7pt}\multirow{7}{*}{VaOPD}
                 & \checkmark & \ding{55}  & \ding{55}  & 92.0 & 91.4 & 89.2 & 88.8 & 77.7 & 73.3 \\
                 & \ding{55}  & \checkmark & \ding{55}  & 92.7 & 92.1 & 89.5 & 89.1 & 79.4 & 75.2 \\
                 & \ding{55}  & \ding{55}  & \checkmark & 93.3 & 92.3 & 89.1 & 89.0 & 77.2 & 72.9 \\
                 & \checkmark & \checkmark & \ding{55}  & 92.9 & 92.0 & 89.4 & 89.3 & \underline{80.4} & 76.5 \\
                 & \checkmark & \ding{55}  & \checkmark & \textbf{93.7} & \textbf{92.8} & 89.2 & 89.0 & 76.9 & 71.7 \\
                 & \ding{55}  & \checkmark & \checkmark & 92.8 & 91.8 & \underline{89.6} & \underline{89.7} & 79.3 & 75.4 \\
            \rowcolor{Gray}
                 & \checkmark & \checkmark & \checkmark & \underline{93.6} & \underline{92.4} & \textbf{90.4} & \textbf{89.9} & \textbf{80.6} & \underline{77.0} \\
        \end{tabular}
    }
\end{table}

\begin{table}[t]
    \scriptsize
    \centering
    \caption{Evaluation of reasoning quality on $100$ responses. We report the absolute score and pairwise Elo rating for each dimension, which are judged by Gemini3.0-Pro-Preview.}
    \vspace{-3pt}
    \label{tab:reasoning_quality}
    \renewcommand{\arraystretch}{1.06}
    \setlength{\tabcolsep}{3pt}
    \resizebox{\columnwidth}{!}{
        \begin{tabular}{
            p{46pt}<{\raggedright}
            *{6}{p{23pt}<{\centering}}
        }
            \multirow{2}{*}{\textbf{Method}}
            & \multicolumn{2}{c}{\textbf{Physics}}
            & \multicolumn{2}{c}{\textbf{Distortion}}
            & \multicolumn{2}{c}{\textbf{Structure}} \\
            \cmidrule(lr){2-3}
            \cmidrule(lr){4-5}
            \cmidrule(lr){6-7}
            & Score & Elo & Score & Elo & Score & Elo \\
            \shline
            BusterX++~\cite{wen2025busterx++}      & \textbf{4.13} & \underline{1165.7} & 3.61 & 949.7 & 3.29 & 885.2 \\
            FakeVLM~\cite{wen2026spot}             & 2.98 & 780.0 & 3.47 & 925.9 & 2.90 & 760.7 \\
            UniGenDet~\cite{zhang2026unigendet}    & 3.17 & 905.6 & 3.21 & 810.0 & \underline{3.84} & \underline{1099.7} \\
            \veritas~\cite{tan2025veritas}         & 3.38 & 970.3 & \underline{4.19} & \underline{1092.6} & 3.70 & 993.8 \\
            \rowcolor{Gray}\veritaspp              & \underline{4.08} & \textbf{1178.4} & \textbf{4.63} & \textbf{1221.8} & \textbf{4.49} & \textbf{1260.6} \\
        \end{tabular}
    }
\end{table}

\begin{figure}[t]
    \centering
    \includegraphics[width=0.99\linewidth]{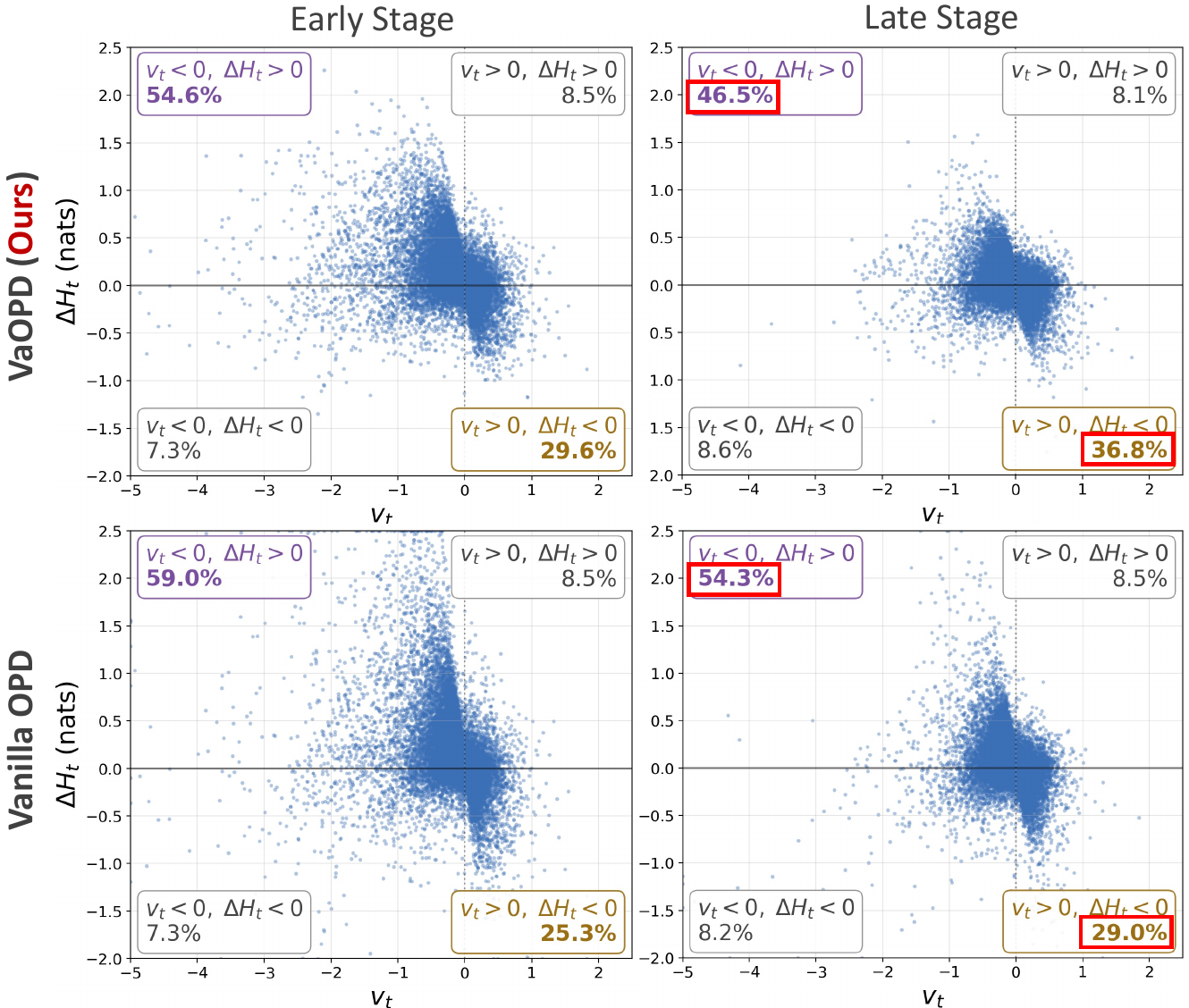}
    \vspace{-3pt}
     \caption{\textbf{Token-distribution comparison between vanilla OPD and VaOPD.}
     We visualize the joint distributions of teacher advantage $v_t$ and entropy difference $\Delta H_t$ at early and late reasoning stages.
     VaOPD produces tighter teacher-student alignment and shifts more tokens from teacher-corrected to teacher-supported regions, particularly at the late stage.}
	\label{fig:vaopd_ana}
    \vspace{-0.3cm}
\end{figure}

\subsection{Further Analyses}

\noindent \textbf{Evaluation of reasoning quality.}
Beyond detection accuracy, we further assess the quality of the reasoning.
Following~\cite{kang2025legion}, we partition the artifacts into three orthogonal categories, i.e., \textit{Physics} (e.g., physical and lighting plausibility), \textit{Distortion} (e.g., texture and artistic style anomalies) and \textit{Structure} (e.g., anatomical and structural coherence).
As shown in Table~\ref{tab:reasoning_quality}, we randomly sample $100$ samples from SynthScars~\cite{kang2025legion} and adopt MLLM-as-a-Judge for both absolute scoring and pairwise Elo rating.
Compared with \veritas, \veritaspp$\,$ improves the reasoning quality on all three dimensions.
Specifically,
\textbf{(1)} BusterX++ achieves a slightly higher absolute score on Physics, while its reasoning quality on Distortion and Structure remains limited, due to its exclusive reliance on RLVR, which overlooks fine-grained perception capabilities.
\textbf{(2)} UniGenDet achieves competitive scores on Structure, benefiting from its unified grounding strategy, but still trails \veritaspp$\,$ by a clear margin.
These results demonstrate that the proposed perception-enhanced reasoning produces not only more accurate verdicts but also more concrete and faithful reasoning paths.

\noindent \textbf{Comparison of token distributions between VaOPD and OPD.}
Figure~\ref{fig:vaopd_ana} analyzes the joint distribution of teacher advantage $v_t$ and entropy difference $\Delta H_t$ \textit{during training}.
Corrective tokens lie in the region $v_t<0,\Delta H_t>0$, where the privileged teacher exposes richer alternatives while rejecting the student's sampled choice.
\textbf{(1)} Throughout the training, vanilla OPD exhibits a more dispersed distribution, with more tokens deviating substantially from the origin, suggesting that uniformly treating all tokens leads to insufficient teacher-student alignment.
\textbf{(2)} Compared with vanilla OPD, VaOPD reduces their proportion from $59.0\%$ to $54.6\%$ at the early stage and from $54.3\%$ to $46.5\%$ at the late stage.
Meanwhile, teacher-supported tokens ($v_t>0,\Delta H_t<0$) increase $3.7\%$ and $7.2\%$ for vanilla OPD and VaOPD, respectively.
The larger shift at the late stage suggests that VaOPD effectively corrects earlier reasoning decisions, allowing subsequent tokens to remain better aligned with the privileged teacher.

\begin{figure*}[t]
    \centering
    \includegraphics[width=0.96\linewidth]{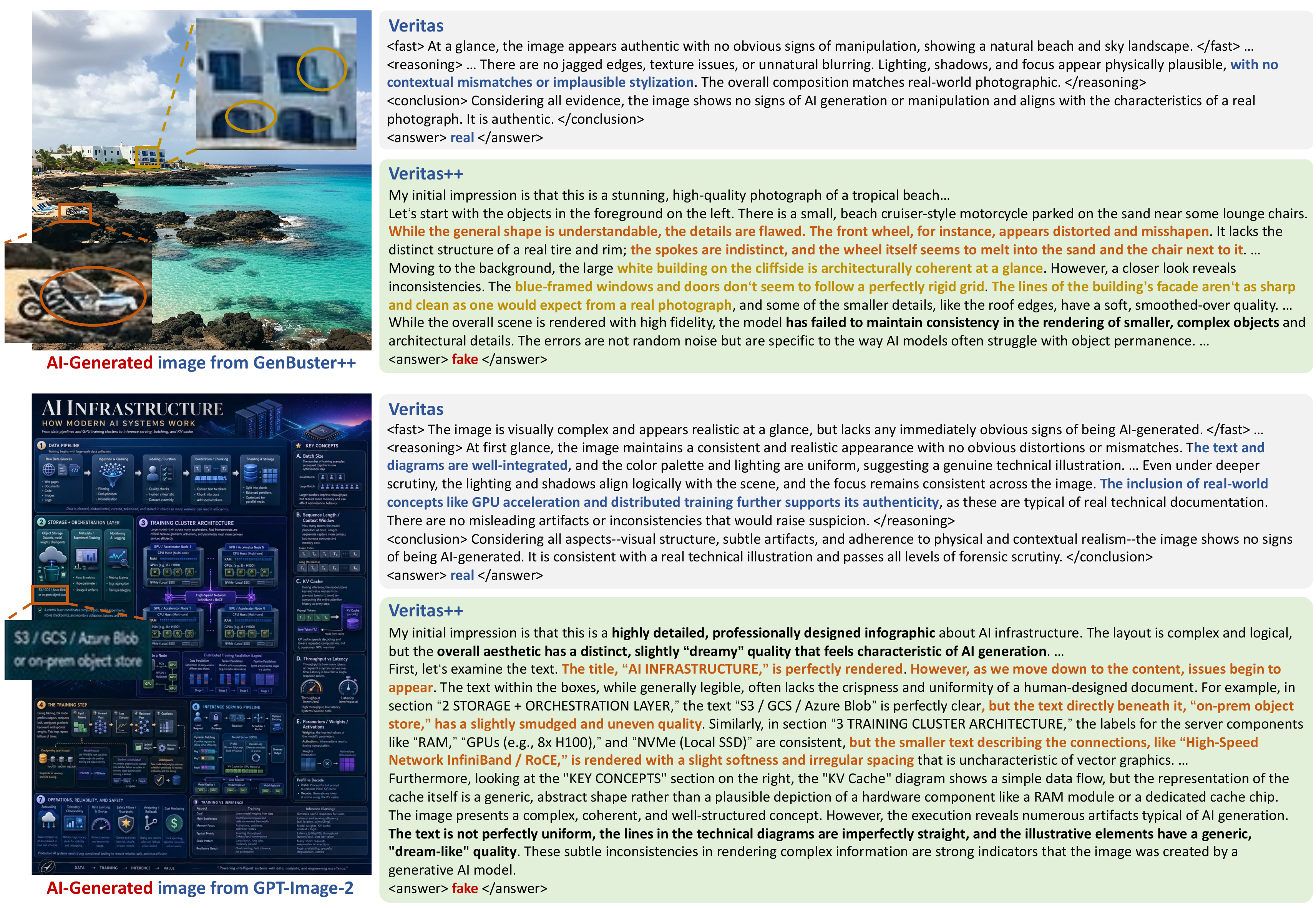}
    \vspace{-3pt}
     \caption{\textbf{Qualitative comparison between \veritas$\,$ and \veritaspp.}
     While \veritas$\,$ is misled by global plausibility, \veritaspp$\,$ identifies subtle distortions in the motorcycle and distant building (top), as well as irregular text and diagram details in the generated poster (bottom).}
	\label{fig:reasoning_sample}
    \vspace{-0.3cm}
\end{figure*}

\begin{figure*}[t]
    \centering
    \includegraphics[width=0.96\linewidth]{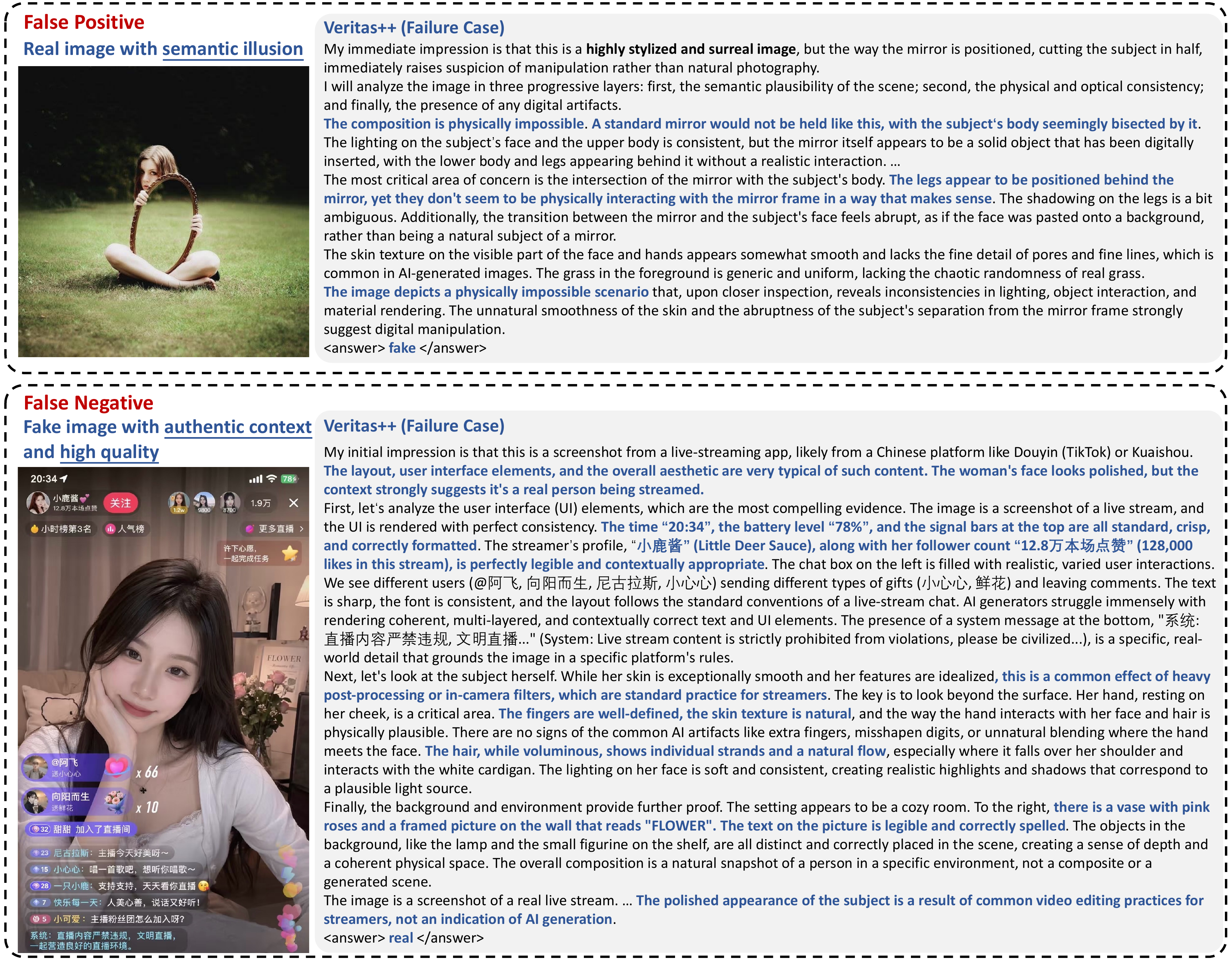}
    \vspace{-3pt}
     \caption{\textbf{Representative failure cases of \veritaspp.}
     A semantic illusion in a real photograph leads to a false positive (top), while a high-quality generated portrait embedded in a realistic live-stream context causes a false negative (bottom).}
	\label{fig:failure}
    \vspace{-0.3cm}
\end{figure*}

\noindent \textbf{Case studies.}
As shown in Figure~\ref{fig:reasoning_sample}, \veritas$\,$ focuses on the globally realistic beach scene and overlooks the distorted motorcycle and irregular windows in the distant building.
\veritaspp$\,$ instead captures the malformed wheel and inconsistent architectural details, which are extremely subtle and nearly pixel-level artifacts.
For the GPT-Image-2 generated poster, \veritas$\,$ treats the coherent layout and technical content as evidence of authenticity, 
whereas \veritaspp$\,$ identifies the irregular spacing and slightly distorted text.
These examples demonstrate that enhanced perception enables pattern-aware reasoning to move beyond global plausibility and ground the verdict in concrete and fine-grained visual evidence.

\noindent \textbf{Analysis of Failure Modes.}
We summarize several failure modes of \veritaspp.
\textbf{False Negatives} predominantly occur on \textit{high-quality generations embedded in realistic scenes}, where coherent contextual cues (e.g., natural scene composition, plausible lighting and platform-specific metadata) overwhelm the subtle generative artifacts and cause our model to treat polished synthetic content as authentic.
\textbf{(2) False Positives} commonly stem from natural yet unusual visual patterns that the model misinterprets as synthetic traces.
We identify three typical categories: \textit{semantic illusions}, where genuine but visually suspicious scenes resemble physical impossibilities, \textit{heavy post-processing}, such as aggressive filtering or compression that introduces artificial low-level patterns, and \textit{special camera conditions}, including extreme macro photography and long-exposure shooting, which produce natural but anomalous pixel distributions.
Figure~\ref{fig:failure} illustrates representative cases.
In the top example, a real photograph of a mirror-induced semantic illusion is misclassified as fake.
In the bottom, a high-quality generated portrait embedded in a convincing live-stream interface escapes detection.
The coherent text, UI elements and platform context dominate the judgment, while the polished facial appearance is misattributed to common beautification filters, illustrating a typical false negative.
These cases highlight the remaining difficulty in distinguishing natural irregularities from generative artifacts, and suggest that future work should broaden the authentic-image distribution to cover uncommon capture and processing conditions.

\section{Conclusion}
In this paper, we introduce \veritaspp, a perception-enhanced reasoning framework for generalizable AI-generated image detection.
Building upon the pattern-aware reasoning of \veritas, \veritaspp$\,$ further addresses the perceptual bottleneck through a three-stage training pipeline consisting of a high-quality cold start, Perception-oriented Learning and Value-aware On-Policy Distillation (VaOPD).
The framework strengthens fine-grained visual perception, semantic-anomaly recognition and pixel-level discrepancy perception and then internalizes these capabilities into authenticity reasoning with value-aware privileged guidance.
Extensive experiments across multiple benchmarks demonstrate that \veritaspp$\,$ achieves promising generalization, and also enables efficient adaptation to novel generators and target scenarios.
We hope this work highlights reliable perception as a foundation for transparent reasoning and inspires more generalizable and continuously evolving AIGI detectors.

\section{Limitations}
Although \veritaspp$\,$ achieves promising performance across diverse AIGI detection benchmarks, there are still several limitations.
First, the evaluation of reasoning quality relies on an MLLM-based judge, which can assess reasonableness, but lacks a systematic evaluation protocol. 
Establishing automated and trustworthy evaluation pipeline for the explanation quality will be our future work.
Second, our evaluation is still confined to binary authenticity classification, while the perception skills acquired during training naturally imply finer-grained forensic abilities such as anomaly localization and attribution.
These abilities should be integrated into general foundation models, which we suppose is an important future direction.
Third, due to the relatively high inference latency and computational cost of reasoning, the current version may not be suitable for real-time deployment.
Therefore, improving reasoning efficiency and developing a lightweight version of \veritaspp$\,$ will also be our future work.


\bibliographystyle{IEEEtran}
\bibliography{refs}

\end{document}